\documentclass[10pt,twocolumn,letterpaper]{article}
\usepackage[margin=0.7in,columnsep=0.25in]{geometry}
\usepackage[T1]{fontenc}
\usepackage[utf8]{inputenc}
\usepackage{lmodern}
\usepackage{amsmath,amssymb,mathtools,bm}
\usepackage{graphicx,booktabs,tabularx,array,makecell,multirow}
\usepackage{caption}
\usepackage{microtype}
\usepackage[numbers,sort&compress]{natbib}
\usepackage{xcolor}
\usepackage[percent]{overpic}
\usepackage{placeins}
\usepackage{stfloats}
\usepackage{needspace}

\usepackage{url}
\usepackage[colorlinks=true,allcolors=blue,breaklinks=true]{hyperref}
\usepackage{bookmark}
\DeclareUnicodeCharacter{221E}{\ensuremath{\infty}}
\DeclareUnicodeCharacter{2192}{\ensuremath{\to}}
\DeclareUnicodeCharacter{2212}{\ensuremath{-}}
\DeclareUnicodeCharacter{2264}{\ensuremath{\le}}
\DeclareUnicodeCharacter{2265}{\ensuremath{\ge}}
\DeclareUnicodeCharacter{00D7}{\ensuremath{\times}}
\hypersetup{pdftitle={HasMem: Hard-Origin Adaptively Softened Memory for Long-Term LLM Agents},
pdfauthor={Zihong He, Junxiao Shen, Chen Liang, Hai-Ning Liang}}
\begin{document}
\twocolumn[{%
\begin{@twocolumnfalse}
\begin{center}
{\LARGE\bfseries HasMem: Hard-Origin Adaptively Softened Memory\\[3pt]
for Long-Term LLM Agents\par}
\vspace{10pt}
{\large Zihong He$^{1}$ \quad Junxiao Shen$^{2}$ \quad Chen Liang$^{1,*}$ \quad Hai-Ning Liang$^{1,3,*}$\par}
\vspace{5pt}
$^{1}$The Hong Kong University of Science and Technology (Guangzhou)\\
$^{2}$University of Bristol\\
$^{3}$The Hong Kong University of Science and Technology\\[3pt]
{\small \texttt{zhe154@connect.hkust-gz.edu.cn}, \texttt{Junxiao.shen@bristol.ac.uk}\\
\texttt{chenliang2@hkust-gz.edu.cn}, \texttt{hainingliang@hkust-gz.edu.cn}\\
$^{*}$Corresponding authors}
\end{center}
\vspace{4pt}
\begin{abstract}Text-based memory and context compression support reuse of past interactions. Resizing continuous memory changes the input to a frozen LLM, coupling capacity allocation with readout. We propose Hard-Origin Adaptively Softened Memory (HasMem). Frozen hard-prompt embeddings provide a verifiable initial state. A controller adjusts memory widths, a Writer re-encodes resized entries, and Reader and Global provide readout adaptation and cross-turn state. On all $535$ questions in a reconstruction probe derived from the Multi-Session Chat (MSC) development split, the main configuration achieves lexical F1 of $95.3$ ($+4.4$ percentage points) at $93.6\%$ of the hard reference's framed memory positions. With approximately matched per-question target body budgets, six configurations at mean per-entry retention around $0.83$--$0.91$ exceed rule-based re-encoding by $8.0$--$23.6$ exact-match (EM) percentage points. With fixed model parameters and rule target width ratio $0.75$, Global's EM gain passes a user-level exact paired test with Bonferroni correction over eight comparisons. On all $500$ LongMemEval-S questions, local lexical F1 rises from the hard reference's $3.4$ to $8.9$, and answer negative log-likelihood (NLL) falls from $12.257$ to $5.274$. F1 gains accompany lower EM on both evaluations.

\end{abstract}
\begin{center}
\small Code: \url{https://github.com/ZihongHe/HasMem}
\end{center}
\vspace{8pt}
\end{@twocolumnfalse}
}]
\section{Introduction}
\label{sec:introduction}

Long-term LLM agents retain information to answer later questions~\citep{zhang2025survey,wu2025human}. Text-based systems manage history as notes or paged context~\citep{chhikara2025mem0,packer2023memgpt}. Continuous compressors produce compact slots~\citep{ge2023context}, and token compressors retain selected text~\citep{pan2024llmlingua}. Adaptive compression selects budgets per input~\citep{zhao2026flexcomp}.

A further challenge for persistent continuous memory is revising stored records' widths, measured in continuous positions, as interactions proceed. Resizing changes the frozen LLM's input, coupling capacity allocation, re-encoding, and readout. The frozen LLM's token embeddings provide a verifiable hard-prompt starting point for gradual per-record width adaptation.

We therefore propose \emph{Hard-Origin Adaptively Softened Memory} (HasMem), which maintains these representations in a \emph{Soft Bank} outside the frozen LLM parameters $W_0$. A Controller chooses KEEP, SHRINK, or EXPAND, and a Writer re-encodes resized entries. KEEP preserves the slot. Reader provides low-rank readout adaptation, and a recurrent Global state conditions re-encoding and readout. A causal gate preserves hard-prompt equivalence for all-KEEP trajectories under identical retrieval and input construction (Section~\ref{sec:inference}). Subsequent question answering supervises training, with a penalty on cumulative memory positions. Evaluation updates use observed history and stored state before receiving the question.

The main configuration uses frozen Qwen2.5-7B, $k=8$. We evaluate all $535$ reconstruction questions derived from the Multi-Session Chat (MSC) development split~\citep{xu2022beyond} and all $500$ LongMemEval-S questions~\citep{wu2024longmemeval}. On LongMemEval-S, local lexical F1 improves by $5.6$ percentage points and answer NLL falls by $6.983$ (Table~\ref{tab:2}). Exact match (EM) is lower than the hard reference on both datasets.

We make three contributions.
\begin{enumerate}
\item \textbf{Continuous memory with a verifiable hard origin.} HasMem initializes records from frozen LLM token embeddings and adapts their widths, preserving all-KEEP hard-prompt equivalence (Section~\ref{sec:inference}). On MSC-derived, the main configuration improves lexical F1 by $4.4$ percentage points at $93.6\%$ of the hard reference's framed positions (Table~\ref{tab:1}).
\item \textbf{Learned width decisions and re-encoding.} A controller selects record widths, and a re-encoding module updates resized entries. Sharing trained parameters and the re-encoder at approximately matched per-question target body budgets, six configurations with mean per-entry retention around $0.83$--$0.91$ exceed rule-based re-encoding by $8.0$--$23.6$ EM percentage points (Table~\ref{tab:4}).
\item \textbf{Readout adaptation with cross-turn context.} We introduce low-rank adaptation (Reader) and recurrent conditioning of re-encoding and readout (Global). With model parameters and per-question positions fixed, Global readout at rule target $\rho=0.75$ raises question-weighted MSC-derived EM by $2.24$ percentage points (owner-level exact paired test~\citep{winkler2014permutation}, Bonferroni-adjusted $p=0.0146$ over eight comparisons~\citep{dunn1961multiple}, Table~\ref{tab:6}).
\end{enumerate}

\section{Related Work}
\label{sec:related}

\subsection{Learning-Based Memory Management}
\label{sec:memory-management}

RAG~\citep{lewis2020retrieval} and RETRO~\citep{borgeaud2022improving} generate from retrieved text. HippoRAG~\citep{gutierrez2024hipporag} uses graph retrieval, and ReadAgent~\citep{lee2024human} combines gist memory with selective rereading. Generative Agents~\citep{park2023generative} retrieves by importance, recency, and relevance. MemoryBank~\citep{zhong2024memorybank} applies time-dependent forgetting. MemGPT~\citep{packer2023memgpt} pages context, while Mem0~\citep{chhikara2025mem0} and A-MEM~\citep{xu2026mem} extract and link notes. Memory-R1~\citep{yan2026memory} learns text-memory operations from task outcomes. Larimar~\citep{das2024larimar} supports one-shot episodic-memory edits. HasMem controls continuous width per record. Appendix~\ref{app:A} evaluates external adaptations with shared data and lexical scoring.

\subsection{Persistent Memory in Continuous Contexts}
\label{sec:continuous-contexts}

Prompt Tuning~\citep{lester2021power} learns continuous input prompts. Prefix-Tuning~\citep{li2021prefix} and P-Tuning v2~\citep{liu2022p} learn multilayer continuous prefixes. Adapters~\citep{houlsby2019parameter} and LoRA~\citep{hu2021lora} add trainable parameters to frozen backbones. Generative Adapter~\citep{chen2025generative} maps context to low-rank adapters. xRAG~\citep{cheng2024xrag} bridges retrieval embeddings to LLMs. Transformer-XL~\citep{dai2019transformer} and Recurrent Memory Transformer~\citep{bulatov2022recurrent} carry states or memory vectors across segments. Compressive Transformer~\citep{rae2019compressive} compresses older states. LongMem~\citep{wang2023augmenting} adds a memory network beside a frozen backbone. The $\infty$-former~\citep{martins2022former} uses continuous-space attention. Memorizing Transformers~\citep{wu2022memorizing} and Focused Transformer~\citep{tworkowski2023focused} retrieve stored keys and values.

Gist Tokens~\citep{mu2023learning}, AutoCompressors~\citep{chevalier2023adapting}, and ICAE~\citep{ge2023context} compress contexts into vectors. MemoryLLM~\citep{wang2024memoryllm} and M+~\citep{wang2025m+} maintain layerwise shared pools. Matryoshka Representation Learning~\citep{kusupati2022matryoshka} supports nested embedding dimensions. FlexComp~\citep{zhao2026flexcomp} trains one compressor across memory-token budgets and predicts write-time budgets from context. HasMem starts each record from hard-prompt embeddings and continually revises its width.

\subsection{Token and Memory Pruning}
\label{sec:pruning}

LLMLingua~\citep{jiang2023llmlingua} and Selective Context~\citep{li2023compressing} prune by perplexity or self-information. LongLLMLingua~\citep{jiang2024longllmlingua} ranks documents and prunes tokens using questions. LLMLingua-2~\citep{pan2024llmlingua} learns task-agnostic extractive compression. Rate--distortion analysis characterizes question-conditioned retention at fixed budgets~\citep{nagle2024fundamental}. SDTP~\citep{tao2025saliency} learns layerwise pruning from gradient saliency. SimpleMem~\citep{liu2026simplemem} compresses conversations into text memories. MemRefine~\citep{kim2026memrefine} compresses memory banks through question-independent deletion, merging, and retention to meet storage budgets. HasMem retains entries and adapts their continuous widths.

StreamingLLM~\citep{xiao2024efficient} retains attention sinks and recent tokens. H$_2$O~\citep{zhang2023h2o} and Scissorhands~\citep{liu2023scissorhands} evict KV entries by attention importance. SnapKV~\citep{li2024snapkv} selects positions from observation-window attention. PyramidKV~\citep{cai2024pyramidkv} allocates across layers, Ada-KV~\citep{feng2026ada} across heads. DMC~\citep{nawrot2024dynamic} learns KV append/merge operations. CCM~\citep{kim2024compressed} recursively compresses states for online interaction. At inference time, HasMem allocates persistent record widths before questions arrive. Table~\ref{tab:4} compares learned trajectories with rule re-encoding at approximately matched per-question target body budgets.

\section{Method}
\label{sec:method}

\begin{figure*}[t]
\centering
\begin{overpic}[width=0.94\textwidth]{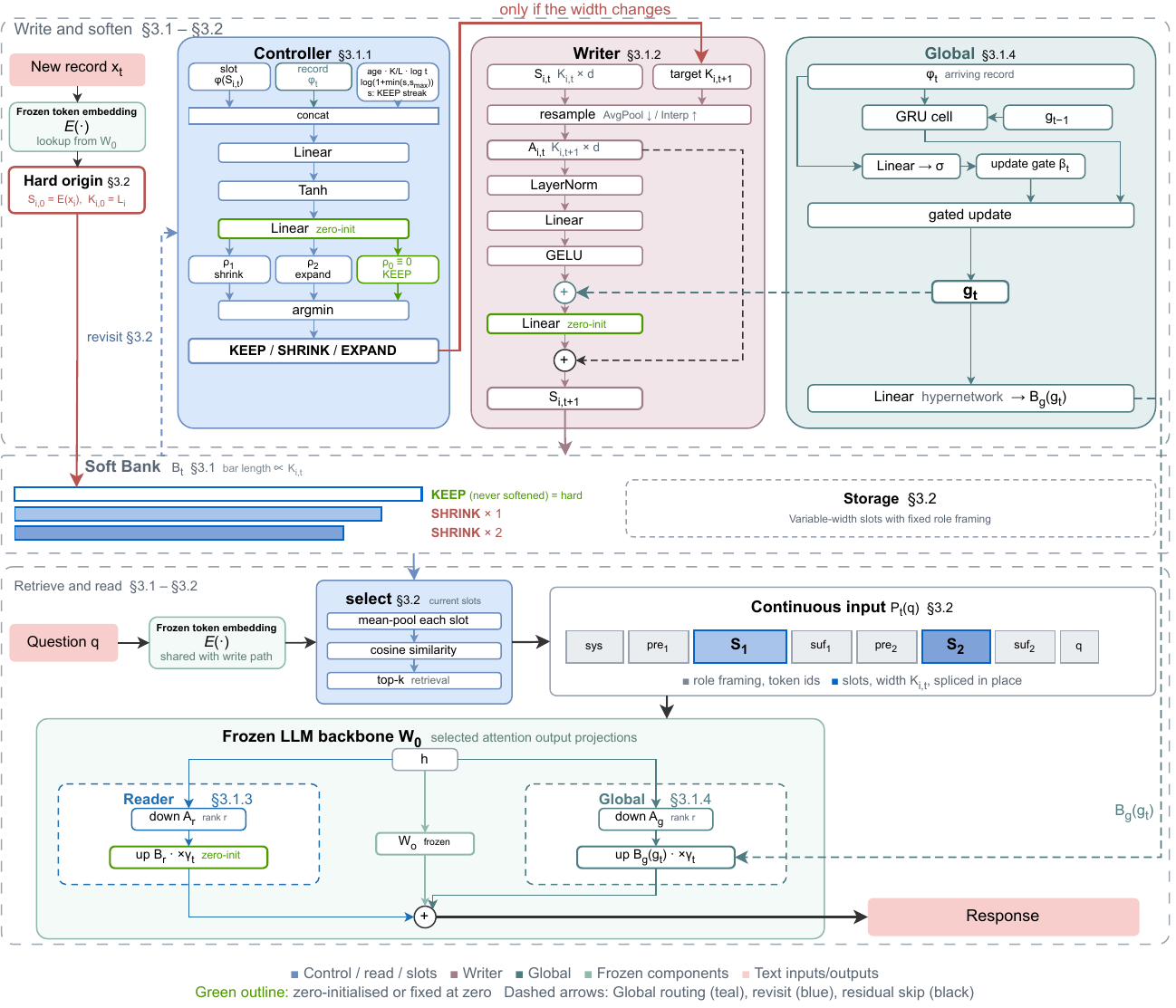}
\put(12.0480,82.4126){\hyperref[sec:architecture]{\phantom{\rule{0.025349\linewidth}{0.014553\linewidth}}}}
\put(16.2857,82.4126){\hyperref[sec:inference]{\phantom{\rule{0.025349\linewidth}{0.014553\linewidth}}}}
\put(16.1985,45.2889){\hyperref[sec:architecture]{\phantom{\rule{0.020068\linewidth}{0.011521\linewidth}}}}
\put(11.0335,35.2299){\hyperref[sec:architecture]{\phantom{\rule{0.021124\linewidth}{0.012127\linewidth}}}}
\put(14.5649,35.2299){\hyperref[sec:inference]{\phantom{\rule{0.021124\linewidth}{0.012127\linewidth}}}}
\put(9.1343,69.6799){\hyperref[sec:inference]{\phantom{\rule{0.017956\linewidth}{0.010308\linewidth}}}}
\put(28.7737,80.4074){\hyperref[sec:controller]{\phantom{\rule{0.027160\linewidth}{0.010915\linewidth}}}}
\put(53.3188,80.4074){\hyperref[sec:writer]{\phantom{\rule{0.027160\linewidth}{0.010915\linewidth}}}}
\put(83.5116,80.4074){\hyperref[sec:global]{\phantom{\rule{0.027160\linewidth}{0.010915\linewidth}}}}
\put(77.1803,42.5962){\hyperref[sec:inference]{\phantom{\rule{0.023237\linewidth}{0.013340\linewidth}}}}
\put(33.7227,34.0995){\hyperref[sec:inference]{\phantom{\rule{0.019012\linewidth}{0.010915\linewidth}}}}
\put(78.8142,33.3822){\hyperref[sec:inference]{\phantom{\rule{0.021124\linewidth}{0.012127\linewidth}}}}
\put(17.9711,16.9594){\hyperref[sec:reader]{\phantom{\rule{0.033195\linewidth}{0.013340\linewidth}}}}
\put(56.9907,16.9594){\hyperref[sec:global]{\phantom{\rule{0.033195\linewidth}{0.013340\linewidth}}}}
\put(10.7437,56.3629){\hyperref[sec:inference]{\phantom{\rule{0.021124\linewidth}{0.012127\linewidth}}}}
\end{overpic}
\caption{HasMem architecture. Records begin as frozen token embeddings and are maintained as variable-width continuous slots. The Controller selects width changes, the Writer re-encodes affected slots, and Reader and Global provide readout increments. Global also maintains a recurrent state across records.}
\label{fig:1}
\end{figure*}
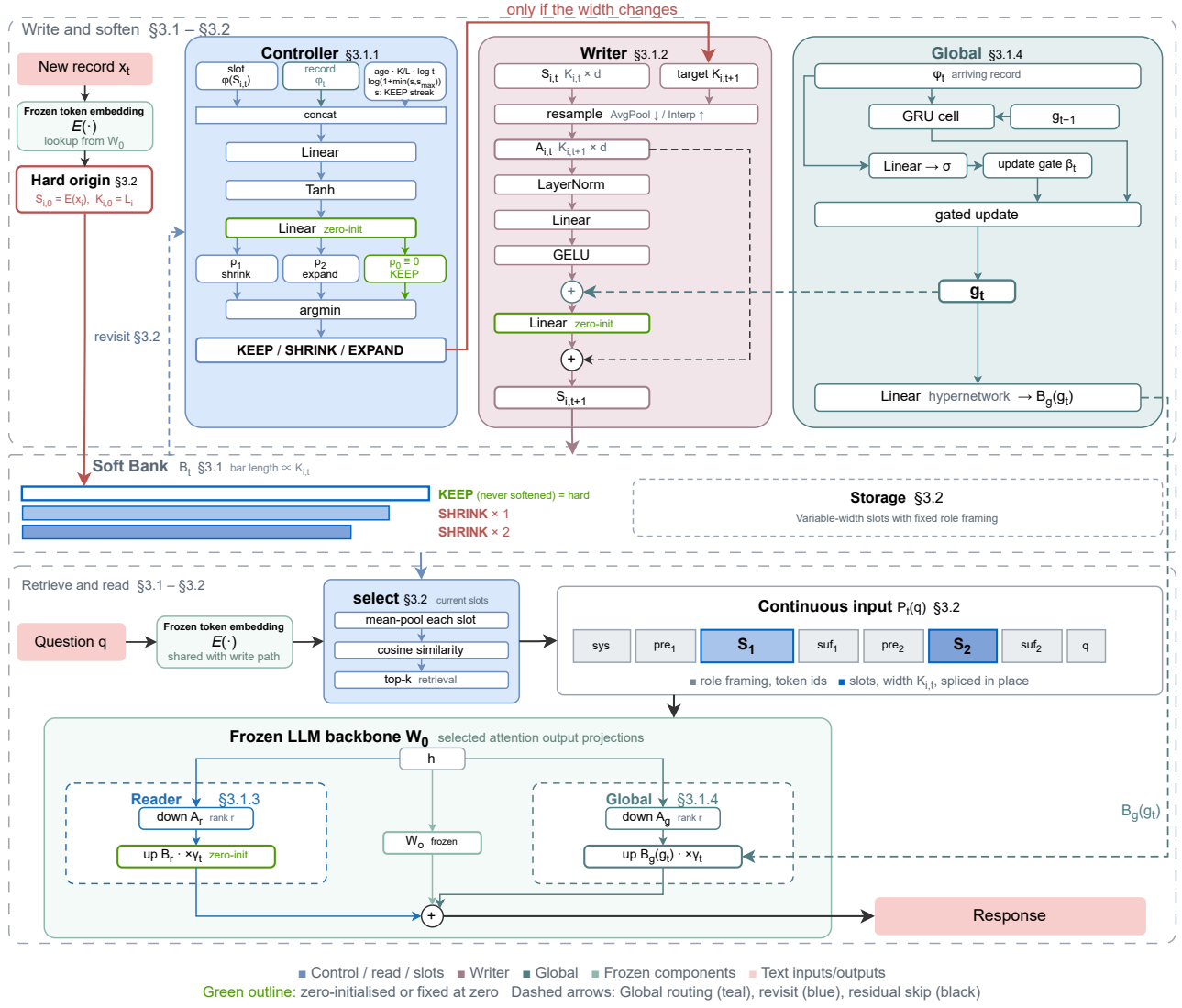

Figure~\ref{fig:1} summarizes HasMem's architecture, inference procedure, and training.

\subsection{Architecture}
\label{sec:architecture}

HasMem maintains a variable-width \emph{Soft Bank} and cross-turn \emph{Global} state outside the frozen backbone $W_0$. For user $u$, step $t=0$ writes the first record. Subsequent records or maintenance increment $t$. In state $\mathcal M_{u,t}=(g_{u,t},\mathcal B_{u,t})$, $g_{u,t}\in\mathbb R^m$ is the Global state. User $u$'s Soft Bank $\mathcal B_{u,t}$ at step $t$ contains $N_{u,t}$ entries in arrival order and is defined by
\begin{equation}
\label{eq:bank}
\mathcal B_{u,t}=\bigl\{(c_i^{\mathrm{pre}},\,S_{i,t},\,c_i^{\mathrm{suf}})\bigr\}_{i=1}^{N_{u,t}}.
\end{equation}
Entry $i$ has a body slot $S_{i,t}\in\mathbb R^{K_{i,t}\times d}$ of width $K_{i,t}$ and backbone hidden dimension $d$, with fixed role-framing token sequences $c_i^{\mathrm{pre}}$ and $c_i^{\mathrm{suf}}$. We omit $u$ where unambiguous.

Controller, Writer, and Global share a position-wise encoder. For a sequence $V=(v_1,\ldots,v_{|V|})$ with $v_j\in\mathbb R^d$, its mean-pooled feature is
\begin{equation}
\label{eq:encoder}
\phi(V)=\frac1{|V|}\sum_{j=1}^{|V|}\mathrm{GELU}\bigl(W_e\,\mathrm{LN}(v_j)\bigr).
\end{equation}
Here $W_e\in\mathbb R^{m\times d}$ is learned. LN and GELU denote layer normalization and the Gaussian error linear unit. Writer retains position-wise features. Equations omit affine biases.

\subsubsection{Controller}
\label{sec:controller}

Controller selects an action from $\mathcal A=\{\mathrm{KEEP},\mathrm{SHRINK},\mathrm{EXPAND}\}$, indexed $0,1,2$, for scheduled entry $i$. The input $z_{i,t}\in\mathbb R^{2m+4}$ combines slot features, the latest record's mean-pooled embedding feature $\phi_t$, and metadata as
\begin{equation}
\label{eq:controller-input}
\begin{aligned}
z_{i,t}=\bigl[\, &\phi(S_{i,t});\ \phi_t;\ \log(1+t-\tau_i);\\
&K_{i,t}/L_i;\ \log(1+t);\\
&\log(1+\min(s_t,s_{\max}))\,\bigr].
\end{aligned}
\end{equation}
Here $L_i$ counts original body tokens. $\tau_i$ is the arrival step, $s_t$ is the preceding KEEP streak, and $s_{\max}$ is its clipping limit. Semicolons concatenate features. Scalars encode age, retention, step, and streak. Maintenance reuses $\phi_t$.

A two-layer multilayer perceptron (MLP) $f_\theta$ with hyperbolic tangent (Tanh) activation predicts costs $(\tilde\rho_1,\tilde\rho_2)=f_\theta(z_{i,t})$ relative to KEEP, with $\tilde\rho_0\equiv0$, giving
\begin{equation}
\label{eq:action}
a_{i,t}=\arg\min_{a\in\{0,1,2\}}\tilde\rho_a.
\end{equation}
The final layer is zero-initialized, and ties select KEEP. For adjustment fraction $\eta\in(0,1)$ and minimum positive integer width $K_{\min}$, set $\Delta K_{i,t}=\lceil\eta K_{i,t}\rceil$, where $\lceil\cdot\rceil$ rounds up. Widths follow
{\small
\begin{equation}
\label{eq:width-update}
\begin{aligned}
K_{i,t+1}&=\begin{cases}
K_{i,t},&\text{KEEP/bounded},\\
\max(K_{\min},K_{i,t}-\Delta K_{i,t}),&\text{SHRINK},\\
\min(L_i,K_{i,t}+\Delta K_{i,t}),&\text{EXPAND}.
\end{cases}
\end{aligned}
\end{equation}
}
SHRINK is bounded when $K_{i,t}\le K_{\min}$. Otherwise the second case applies. Unchanged widths count as KEEP. Widths for records with $L_i\ge K_{\min}$ stay within $[K_{\min},L_i]$. Shorter records retain their initial width.

\subsubsection{Soft Bank and Writer}
\label{sec:writer}

Writer alone changes slot contents when $K_{i,t+1}\ne K_{i,t}$, resampling positions independently per hidden channel using
\begin{equation}
\label{eq:resampling}
A_{i,t}=\begin{cases}
\mathrm{AvgPool}(S_{i,t};K_{i,t+1}),&K_{i,t+1}<K_{i,t},\\
\mathrm{Interp}(S_{i,t};K_{i,t+1}),&K_{i,t+1}>K_{i,t}.
\end{cases}
\end{equation}
AvgPool and Interp denote adaptive average pooling and linear interpolation to the second argument's target width. A learned up-projection $U\in\mathbb R^{d\times m}$ adds a Global-conditioned residual at position $j$, yielding
\begin{equation}
\label{eq:writer}
\begin{aligned}
S_{i,t+1}[j]=A_{i,t}[j]+U\Bigl(&\mathrm{GELU}\bigl(W_e\mathrm{LN}(A_{i,t}[j])\bigr)\\
&+g_t\Bigr).
\end{aligned}
\end{equation}
$U$ is zero-initialized. KEEP preserves slots. SHRINK and EXPAND re-encode current states with $g_t$. EXPAND is bounded by $L_i$.

\subsubsection{Reader}
\label{sec:reader}

Question-answering (QA) supervision trains Reader to read softened slots through low-rank adaptation~\citep{hu2021lora} of selected attention output projections. For layer $\ell$ with frozen projection $W_o^{(\ell)}$ and input $h\in\mathbb R^d$, Reader adds
\begin{equation}
\label{eq:reader}
\Delta_{\mathrm R}^{(\ell)}(h)=B_r^{(\ell)}A_r^{(\ell)}h.
\end{equation}
Learned down- and up-projections $A_r^{(\ell)}\in\mathbb R^{r\times d}$ and $B_r^{(\ell)}\in\mathbb R^{d\times r}$ use adaptation rank $r$. $B_r^{(\ell)}$ is zero-initialized. Parameters remain fixed at inference.

\subsubsection{Global}
\label{sec:global}

A gated recurrent unit (GRU) updates Global in record order from $g_{-1}=\mathbf0$, gated by $\beta_t=\sigma(w_r^\top\phi_t+b_r)$ with sigmoid function $\sigma$, learned weight $w_r\in\mathbb R^m$, and bias $b_r$. The update is
\begin{equation}
\label{eq:global-update}
g_t=g_{t-1}+\beta_t\bigl(\mathrm{GRU}(\phi_t,g_{t-1})-g_{t-1}\bigr).
\end{equation}
$w_r$ is zero-initialized. Maintenance leaves $g_t=g_{t-1}$. In Reader's attention layers, an affine hypernetwork $B_g^{(\ell)}:\mathbb R^m\to\mathbb R^{d\times r}$ generates a state-conditioned up-projection used in
\begin{equation}
\label{eq:global-readout}
\Delta_{\mathrm G}^{(\ell)}(h;g_t)=B_g^{(\ell)}(g_t)A_g^{(\ell)}h.
\end{equation}
The down-projection $A_g^{(\ell)}\in\mathbb R^{r\times d}$ is learned. Hypernetwork parameters are zero-initialized and fixed after training. $g_t$ conditions the generated projection and Writer.

\subsection{Inference Procedure}
\label{sec:inference}

\paragraph{Writing and hard initialization.}
The tokenizer and chat template separate body tokens $x_i$ of length $L_i=|x_i|$ from role framing. With frozen embedding lookup $E(\cdot)$, record features update Global, after which the new slot is initialized as
\begin{equation}
\label{eq:hard-init}
S_{i,0}=E(x_i)\in\mathbb R^{L_i\times d}.
\end{equation}
Subscript $0$ denotes local initialization, with $K_{i,0}=L_i$. Inspired by input-level continuous prompts~\citep{lester2021power}, slots preserve token order, role boundaries, and full width on chronological arrival.

\paragraph{Softening.}
Each new-record step visits at most one entry. With positive-integer round-robin interval $c_{\mathrm{rev}}$ and $t>0$, its zero-based index is $j_t=\lfloor(t-1)/c_{\mathrm{rev}}\rfloor\bmod N_t$. Preset post-history maintenance continues this schedule without adding records or updating Global. Arrival order and maintenance duration determine visit counts.

\paragraph{Retrieval.}
For question tokens $q$, entry $i$ of width $K_{i,t}$, and $d$-dimensional position vectors $S_{i,t}[j]$ and $E(q)[j]$, the retrieval score is the cosine similarity of position-wise means
\begin{equation}
\label{eq:retrieval}
\mathrm{sim}(i,q)=\cos\left(\frac1{K_{i,t}}\sum_{j=1}^{K_{i,t}}S_{i,t}[j],\frac1{|q|}\sum_{j=1}^{|q|}E(q)[j]\right),
\end{equation}
Let $\mathcal R_k(q)$ index the top-scoring, at most $k$ entries in arrival order. With system-prompt tokens $c^{\mathrm{sys}}$ and chat-templated question tokens $c^q$, the input is
\begin{equation}
\label{eq:prompt}
\begin{aligned}
P_t(q)=\bigl[&E(c^{\mathrm{sys}});\\
&\{E(c_i^{\mathrm{pre}}),S_{i,t},E(c_i^{\mathrm{suf}})\}_{i\in\mathcal R_k(q)};E(c^q)\bigr].
\end{aligned}
\end{equation}
Semicolons concatenate positions. Slot-derived retrieval keys allow softening to change the selected set. Compared strategies share retrieval and $k$. Paired evaluations record realized sets and budgets.

\paragraph{State-conditioned readout.}
Readout increments activate after the first width change through the gate
\begin{equation}
\label{eq:causal-gate}
\gamma_t=\mathbf{1}\bigl[\exists i,\ 0\le s\le t:\ a_{i,s}\ne\mathrm{KEEP}\bigr].
\end{equation}
The indicator $\mathbf{1}[\cdot]$ equals $1$ when its condition holds. $s$ indexes preceding steps. Unchanged-width actions count as KEEP, giving the effective projection
\begin{equation}
\label{eq:effective-projection}
\begin{aligned}
\widetilde W_o^{(\ell)}(g_t)
&=W_o^{(\ell)}+\gamma_tB_r^{(\ell)}A_r^{(\ell)}\\
&\quad+\gamma_tB_g^{(\ell)}(g_t)A_g^{(\ell)}.
\end{aligned}
\end{equation}
At inference, parameters are fixed and $g_t$ evolves.

\paragraph{Hard equivalence.}
An all-KEEP trajectory preserves $S_{i,t}=E(x_i)$ and $\gamma_t=0$. Under identical retrieval, ordering, tokenization, chat templates, positional encoding, attention masks, and inference settings, HasMem and the hard reference have identical inputs, frozen operators, and logits.

\paragraph{Position budgets.}
Total bank positions, including role framing, are
\begin{equation}
\label{eq:position-budget}
C_t=\sum_{i=1}^{N_t}\bigl(K_{i,t}+|c_i^{\mathrm{pre}}|+|c_i^{\mathrm{suf}}|\bigr).
\end{equation}
Framing lengths count tokens. Top-$k$ input length depends on retrieved entries. $R_{\mathrm{all}}$ is the bank total divided by the hard reference's total. Each table specifies aggregation. With $b_s,b_g,b_c$ bytes per slot-vector, Global-state, and framing-ID element, persistent numerical payload is
\begin{equation}
\label{eq:payload}
b_sd\sum_{i=1}^{N_t}K_{i,t}+b_gm+b_c\sum_{i=1}^{N_t}\bigl(|c_i^{\mathrm{pre}}|+|c_i^{\mathrm{suf}}|\bigr).
\end{equation}
Storage also includes metadata, containers, and module parameters. Position ratios measure representation length. Latency and GPU memory depend on implementation.

For Table~\ref{tab:4}, $\mathcal R_{\mathrm{adaptive}}(q)$ and $\mathcal R_{\mathrm{hard}}(q)$ are the retrieval sets of HasMem and the hard reference. $K_i$ is answer-time width. The target body ratio is
\begin{equation}
\label{eq:body-ratio}
R_{\mathrm{body}}(q)=\frac{\sum_{i\in\mathcal R_{\mathrm{adaptive}}(q)}K_i}{\sum_{i\in\mathcal R_{\mathrm{hard}}(q)}L_i}.
\end{equation}
Equally weighting evaluation questions $\mathcal Q_{\mathrm{eval}}$ and their retrieved entries $\mathcal R(q)$ gives the mean per-entry retention $\bar r$
\begin{equation}
\label{eq:retention}
\bar r=\frac1{|\mathcal Q_{\mathrm{eval}}|}\sum_{q\in\mathcal Q_{\mathrm{eval}}}\frac1{|\mathcal R(q)|}\sum_{i\in\mathcal R(q)}\frac{K_i}{L_i}.
\end{equation}

\subsection{Training}
\label{sec:training}

Training uses QA supervision from each user's (owner's) history, with $W_0$ frozen. Warm-up updates all added modules. Policy training updates Controller, the shared encoder, and Writer's output layer. Reader- and Global-specific parameters retain warm-up values while Global state evolves.

\paragraph{Quality and position cost.}
For an owner's training questions $Q$, let $\ell_{a,q}$ and $\ell_{h,q}$ be answer NLLs for action branch $a$ and the hard reference. Weighting NLL increases by $\lambda_{\mathrm{damage}}$ gives the quality objective
\begin{equation}
\label{eq:quality}
Q_a=\frac1{|Q|}\sum_{q\in Q}\left(\ell_{a,q}+\lambda_{\mathrm{damage}}[\ell_{a,q}-\ell_{h,q}]_+\right),
\end{equation}
Here $[x]_+=\max(x,0)$, and $c_a$ is cumulative framed bank positions across writing and maintenance divided by the hard reference's rollout total. With position-cost weight $\lambda_{\mathrm{cost}}$, the branch score is
\begin{equation}
\label{eq:branch-score}
J_a=Q_a+\lambda_{\mathrm{cost}}c_a.
\end{equation}
Warm-up minimizes mean $Q_a$ across KEEP, SHRINK, and EXPAND and trains Controller on stopped-gradient regression targets $(J_1-J_0,J_2-J_0)$.

\paragraph{Policy supervision.}
Bounded search on training QA finds sequences containing SHRINK that reduce cumulative positions, meet per-question NLL tolerance against matched KEEP, and improve the search objective. The best sequence supplies its first non-KEEP label. Without a qualifying sequence, a scheduled probability permits a cost-reducing SHRINK weak label, with KEEP used when no weak label is obtained. Appendix~\ref{app:F} specifies search and weighting.

For input $z=z_{i,t}$ and predicted costs $\tilde\rho=(0,\tilde\rho_1,\tilde\rho_2)$, action probabilities are
\begin{equation}
\label{eq:policy}
\pi_\theta(a\mid z)=\operatorname{softmax}(-\tilde\rho)_a.
\end{equation}
Combining weighted action classification $\mathcal L_\pi$, a non-KEEP frequency penalty $\mathcal L_{\mathrm{band}}$ outside a scheduled target band, Writer's answer-NLL damage hinge $\mathcal L_{\mathrm{hinge}}$, and a KEEP-streak penalty $\mathcal L_{\mathrm{streak}}$ on qualifying positives gives
\begin{equation}
\label{eq:policy-loss}
\mathcal L_{\mathrm{policy}}=\mathcal L_\pi+\mathcal L_{\mathrm{band}}+\mathcal L_{\mathrm{hinge}}+\mathcal L_{\mathrm{streak}}.
\end{equation}
Losses average equally over owners. Appendix~\ref{app:F} defines the losses, curriculum, and supervision. Section~\ref{sec:setup} gives experimental settings.

\section{Experiments}
\label{sec:experiments}

\begin{table*}[t]
\centering
\caption{Reconstruction and position cost on MSC-derived 535 (Qwen2.5-7B). Cells give HasMem / hard reference. F1 and EM are percentages. $R_{\mathrm{all}}$ includes role framing.}
\label{tab:1}
\small
\setlength{\tabcolsep}{5pt}
\begin{tabular}{lccccc}
\toprule
Backbone & $k$ & \makecell{F1 $\uparrow$\\HasMem / hard} & \makecell{EM $\uparrow$\\HasMem / hard} & \makecell{NLL $\downarrow$\\HasMem / hard} & \makecell{$R_{\mathrm{all}}\downarrow$\\HasMem / hard} \\
\midrule
Qwen2.5-7B & 2 & 86.7 / 87.6 & 69.6 / 84.7 & 0.463 / 0.416 & 0.900 / 1.000 \\
Qwen2.5-7B & 6 & 95.4 / 91.0 & 65.7 / 88.2 & 0.154 / 0.074 & 0.879 / 1.000 \\
Qwen2.5-7B & 8 & 95.3 / 90.9 & 84.3 / 88.1 & 0.115 / 0.074 & 0.936 / 1.000 \\
Qwen2.5-7B & 12 & 92.3 / 90.9 & 77.4 / 88.1 & 0.173 / 0.074 & 0.921 / 1.000 \\
\bottomrule
\end{tabular}
\end{table*}

\begin{table*}[t]
\centering
\caption{Answer quality on all 500 LongMemEval-S questions. Cells give HasMem / hard reference. Metrics except NLL are percentages. The local judge is Qwen-judge. Training settings appear in Table~\ref{tab:E1}.}
\label{tab:2}
\footnotesize
\setlength{\tabcolsep}{3pt}
\begin{tabular}{lccccccc}
\toprule
Backbone & $k$ & \makecell{F1 $\uparrow$\\HasMem / hard} & \makecell{EM $\uparrow$\\HasMem / hard} & \makecell{NLL $\downarrow$\\HasMem / hard} & \makecell{Cover $\uparrow$\\HasMem / hard} & \makecell{Brief $\uparrow$\\HasMem / hard} & \makecell{Local judge $\uparrow$\\HasMem / hard} \\
\midrule
Qwen2.5-0.5B & 8 & 5.4 / 5.6 & 0.00 / 1.00 & 4.663 / 3.649 & 8.4 / 8.8 & 0.6 / 8.4 & 10.8 / 10.6 \\
Qwen2.5-1.5B & 4 & 7.0 / 3.2 & 0.80 / 1.20 & 5.058 / 4.962 & 9.8 / 7.4 & 3.4 / 7.4 & 8.6 / 7.6 \\
Qwen2.5-1.5B & 8 & 1.3 / 3.6 & 0.40 / 1.40 & 5.931 / 4.911 & 1.4 / 8.2 & 1.0 / 8.2 & 5.0 / 8.4 \\
Qwen2.5-3B & 8 & 6.8 / 3.6 & 0.00 / 0.20 & 4.842 / 10.413 & 15.4 / 10.0 & 0.4 / 7.6 & 9.8 / 10.6 \\
Qwen2.5-7B & 2 & 0.5 / 1.2 & 0.00 / 0.20 & 8.805 / 13.200 & 0.0 / 7.6 & 0.0 / 7.0 & 2.2 / 7.6 \\
Qwen2.5-7B & 4 & 2.0 / 2.7 & 0.00 / 1.20 & 5.328 / 12.985 & 5.4 / 9.0 & 2.0 / 8.4 & 3.8 / 9.2 \\
Qwen2.5-7B & 6 & 8.8 / 3.1 & 1.00 / 1.40 & 4.759 / 12.536 & 12.4 / 10.2 & 5.8 / 9.0 & 11.6 / 10.8 \\
Qwen2.5-7B & 8 & 8.9 / 3.4 & 0.60 / 1.80 & 5.274 / 12.257 & 14.2 / 10.6 & 5.6 / 9.8 & 13.4 / 10.8 \\
Qwen2.5-7B & 12 & 7.5 / 4.7 & 0.00 / 2.40 & 5.491 / 11.534 & 16.4 / 12.8 & 3.4 / 10.2 & 12.6 / 13.0 \\
Qwen2.5-14B & 8 & 8.0 / 2.8 & 0.00 / 0.00 & 6.413 / 17.193 & 16.2 / 12.6 & 3.0 / 9.4 & 12.4 / 13.2 \\
Mistral-7B & 8 & 8.1 / 8.1 & 1.00 / 0.80 & 4.029 / 5.543 & 8.2 / 13.0 & 5.0 / 5.2 & 12.0 / 12.8 \\
\bottomrule
\end{tabular}
\end{table*}
\subsection{Experimental Setup}
\label{sec:setup}
\label{sec:experimental-setup}

\paragraph{Implementation and comparisons.}
The default frozen backbone is Qwen2.5-7B-Instruct~\citep{qwen2025qwen25technicalreport}. Cross-family runs use Mistral-7B-Instruct~\citep{jiang2023mistral7b}. Training uses field QA from the MSC training split and an approximately six-hour wall-clock budget, followed by external evaluation without further training. Tables~\ref{tab:G1} and~\ref{tab:E1} give defaults and actual update counts. Tables~\ref{tab:1}--\ref{tab:2} use the full curriculum and seed $2026091331$, with $k=8$ for the main run. Each hard reference preserves unsoftened token embeddings and shares HasMem's backbone, records, retrieval function, and maximum retrieved-entry count $k$. LongMemEval-S covers all $11$ configurations. MSC-derived covers the trained 7B models at $k=2,6,8,12$. The notation \texttt{full} denotes the complete curriculum and modules at $k=8$, and \texttt{full}$-X$ specifies a component change.

\paragraph{MSC-derived reconstruction probe.}
MSC supplies multi-session conversations and speaker profiles~\citep{xu2022beyond}. To assess retention and reconstruction of written information, deterministic templates convert profile and dialogue values into tagged records. Questions identify tags and require the stored values as answers. Preparation preserves user grouping and record order (filters and target selection in Appendix~\ref{app:G2}). Filtering the $272$ development owners yields $268$ owners, $646$ events, and $535$ questions, with two questions per owner except one and at most $7$ records per owner. Thus $k=8$ retrieves all records. Tag lookup and value copying attain F1/EM of $100.0$. Configuration and checkpoint analyses use this development probe. The test split remains held out.

\paragraph{External evaluation.}
LongMemEval-S's $500$ questions test cross-session factual recall, knowledge updates, temporal reasoning, and information synthesis~\citep{wu2024longmemeval}. Memory is maintained per owner for MSC-derived and independently per question for LongMemEval-S. External-component protocols appear in Appendix~\ref{app:A}.

\paragraph{Metrics.}
Lexical F1 and EM use SQuAD normalization~\citep{rajpurkar2016squad}. MSC-derived EM requires the complete normalized record value. Teacher-forced answer NLL averages negative natural log-probabilities over gold tokens, excluding the end token. Within each backbone, paired scores share gold tokens and scoring positions. Cross-backbone NLL depends on tokenization and vocabulary. MSC-derived scores are owner-weighted unless specified. LongMemEval-S scores are question-weighted. Differences and ratios use unrounded values. Cover applies local answer-coverage rules. Brief additionally limits normalized prediction length to $\max(12,3n_{\mathrm{gold}})$ words, where $n_{\mathrm{gold}}$ counts normalized gold-answer words. Qwen-judge consistently uses official LongMemEval yes/no prompts with local Qwen2.5-7B-Instruct. Scoring details appear in Appendix~\ref{app:G2}.

\paragraph{Paired inference and training variation.}
Fixed-model readout effects are question-weighted EM differences. Treating owners as independent units, two-sided exact tests flip all paired outcomes within each owner~\citep{winkler2014permutation}. Intervals use owner-clustered paired bootstrap~\citep{field2007bootstrapping}. The eight rule-width and two learned-trajectory comparisons form separate Bonferroni families (adjusted $p<0.05$)~\citep{dunn1961multiple}. Pointwise $95\%$ bootstrap intervals are unadjusted. Appendix~\ref{app:G2} gives calculation details. Eight fixed seeds with common training budgets and checkpoint selection (Table~\ref{tab:C1}) capture variation in training, selection, and realized budgets.

\subsection{Results}
\label{sec:results}

\begin{table*}[t]
\centering
\caption{Target-band penalty sweep (MSC-derived 535, $k=8$). Cells give owner-weighted EM (percent) / $R_{\mathrm{all}}$. Single-seed runs target $5465$ updates. 1.5B at $\lambda_{\mathrm{band}}=0$ and 14B at $0.6$ complete $5132$ and $4386$, respectively. Full metrics appear in Table~\ref{tab:B1}.}
\label{tab:3}
\small
\setlength{\tabcolsep}{5pt}
\begin{tabular}{lccccc}
\toprule
$\lambda_{\mathrm{band}}$ & \makecell{0.5B\\EM $\uparrow$ / $R_{\mathrm{all}}\downarrow$} & \makecell{1.5B\\EM $\uparrow$ / $R_{\mathrm{all}}\downarrow$} & \makecell{3B\\EM $\uparrow$ / $R_{\mathrm{all}}\downarrow$} & \makecell{7B\\EM $\uparrow$ / $R_{\mathrm{all}}\downarrow$} & \makecell{14B\\EM $\uparrow$ / $R_{\mathrm{all}}\downarrow$} \\
\midrule
$0.0$ & 12.3 / 0.670 & 9.0 / 0.602 & 77.2 / 0.889 & 75.2 / 0.887 & 45.1 / 0.814 \\
$0.6$ (default) & 7.1 / 0.604 & 17.0 / 0.665 & 72.0 / 0.892 & 38.4 / 0.788 & 49.8 / 0.814 \\
$1.5$ & 9.9 / 0.632 & 28.2 / 0.714 & 77.8 / 0.915 & 80.0 / 0.917 & 84.3 / 0.916 \\
$3.0$ & 9.9 / 0.637 & 49.4 / 0.909 & 57.8 / 0.909 & 90.5 / 0.992 & 93.3 / 1.000 \\
$6.0$ & 63.2 / 0.924 & 79.9 / 0.957 & 80.6 / 0.973 & 86.0 / 0.970 & 91.8 / 0.974 \\
Hard reference & 51.1 / 1.000 & 80.8 / 1.000 & 74.6 / 1.000 & 88.1 / 1.000 & 93.3 / 1.000 \\
\bottomrule
\end{tabular}
\end{table*}

\begin{table*}[!t]
\centering
\caption{Learned versus rule-based re-encoding at approximately matched per-question target body budgets (MSC-derived 535, frozen Qwen2.5-7B, $k=8$ unless specified). Scores are question-weighted. The hard reference has F1/EM of $90.9/88.0$. Rows follow learned retention $\bar r$. The main run is Table~\ref{tab:1}'s seed $2026091331$, final update $5465$, with a wall-clock curriculum. The position gap is learned minus rule-based body positions as a percentage of the hard reference's body positions (question mean / maximum). $\Delta$EM is learned minus rule-based, before rounding.}
\label{tab:4}
\footnotesize
\setlength{\tabcolsep}{3pt}
\begin{tabular}{>{\raggedright\arraybackslash}p{3.55cm}cccccc}
\toprule
Trained configuration & \makecell{Learned\\$\bar r$} & \makecell{Learned\\F1 / EM} & \makecell{Rule-based\\$\bar r$} & \makecell{Rule-based\\F1 / EM} & \makecell{$\Delta$EM\\learned $-$ rule-based} & \makecell{Position gap (\%)\\mean / maximum} \\
\midrule
Seed 2026091703 & $0.952$ & 93.7 / 79.6 & $0.952$ & 93.8 / 79.8 & $-0.2$ & 0.05 / 2.40 \\
Seed 2026091401 & $0.946$ & 96.1 / 82.2 & $0.946$ & 96.5 / 82.6 & $-0.4$ & 0.05 / 2.40 \\
Main run & $0.911$ & 95.5 / 84.5 & $0.910$ & 93.9 / 66.5 & $+17.9$ & 1.06 / 5.41 \\
Seed 2026091702 & $0.884$ & 93.6 / 79.3 & $0.883$ & 94.3 / 65.4 & $+13.8$ & 1.35 / 5.95 \\
Seed 2026091701 & $0.883$ & 93.7 / 77.6 & $0.882$ & 92.1 / 54.0 & $+23.6$ & 1.22 / 5.41 \\
\texttt{full} (fixed steps) & $0.847$ & 94.1 / 75.0 & $0.841$ & 93.9 / 55.5 & $+19.4$ & 1.64 / 5.13 \\
\texttt{full}, $k=6$ & $0.840$ & 95.5 / 65.8 & $0.838$ & 94.7 / 56.6 & $+9.2$ & 0.68 / 4.69 \\
Gaussian-input-trained & $0.833$ & 93.7 / 58.5 & $0.832$ & 88.8 / 50.5 & $+8.0$ & 0.60 / 4.69 \\
\texttt{full}, width step $20\%$ & $0.765$ & 87.2 / 39.1 & $0.760$ & 88.8 / 40.6 & $-1.5$ & 1.20 / 4.48 \\
\texttt{full} $-$ target band & $0.761$ & 91.9 / 48.0 & $0.759$ & 90.2 / 44.5 & $+3.6$ & 0.84 / 4.11 \\
\texttt{full}, final target $30\%$ & $0.745$ & 83.9 / 34.6 & $0.741$ & 82.5 / 31.6 & $+3.0$ & 0.41 / 4.29 \\
\texttt{full} $-$ forced softening & $0.683$ & 84.0 / 30.7 & $0.678$ & 81.6 / 32.9 & $-2.2$ & 0.99 / 4.00 \\
\bottomrule
\end{tabular}
\end{table*}

\subsubsection{Comparison with Hard-Prompt References}
\label{sec:hard-comparison}

\paragraph{Reconstruction and position cost.}
The main configuration uses $93.6\%$ of the hard reference's framed positions ($R_{\mathrm{body}}\approx0.921$, $\bar r\approx0.911$), with F1 $4.4$ percentage points higher, EM $3.7$ percentage points lower, and NLL $0.041$ higher. Under identical instructions, the hard reference and HasMem output UNKNOWN on $34/12$ questions. Their answered subsets have question-weighted F1 $97.0/97.7$ and EM $94.0/86.4$. F1 gains accompany fewer abstentions and lower sentence agreement.

\paragraph{Cross-session QA.}
On LongMemEval-S, the main configuration improves F1/Cover by $5.6/3.6$ percentage points and reduces NLL by $6.983$. Qwen-judge scores are $13.4$ versus the hard reference's $10.8$. Brief/EM decline by $4.2/1.2$ percentage points. Answers satisfying coverage while exceeding the length limit constitute $8.6\%$ of questions for HasMem versus $0.8\%$ for the hard reference.

\paragraph{Configuration dependence.}
NLL improves in eight of eleven configurations. Local judge scores improve in four and decline in seven. MSC-derived EM decreases at every retrieval width. Runs complete $2907$--$5879$ updates within the common wall-clock budget (Table~\ref{tab:E1}). Comparisons therefore reflect backbone, retrieval width, and update-count differences.

\Needspace{5\baselineskip}
\subsubsection{Configuration and Strategy Comparisons}
\label{sec:configuration-comparison}
\paragraph{Target-band penalty.}
Table~\ref{tab:3} varies $\lambda_{\mathrm{band}}\in\{0,0.6,1.5,3,6\}$ across five Qwen2.5 backbones with other planned settings fixed (Appendix~\ref{app:G}). Each backbone's highest EM occurs at $\lambda_{\mathrm{band}}\ge3.0$ and relative budgets $0.924$--$1.000$. EM is nonmonotonic for four backbones. The 14B model at $\lambda_{\mathrm{band}}=3.0$ selects KEEP throughout and matches the hard reference.

\paragraph{Training variation.}
Eight seeds yield mean F1/EM $94.2/82.6$ and EM range $77.4$--$88.1$ (Appendix~\ref{app:C}). Tables~\ref{tab:E2}--\ref{tab:E4} cover hyperparameters, action policies, and training recipes. Under common deployment, all four recipes increase LongMemEval-S F1 and decrease its NLL and MSC-derived EM relative to the hard reference. Objectives, input distributions, and realized budgets vary jointly.

\paragraph{Width trajectories at approximately matched target budgets.}
Table~\ref{tab:4} shares trained parameters, Writer, and hard initialization across strategies. The rule shrinks to each learned trajectory's target $R_{\mathrm{body}}$ or below, leaving discrete residuals. Learned re-encoding occurs during writing and maintenance. Rule-based re-encoding follows the completed history. Twelve configurations span retention levels, including settings selected for compression depth. All eleven $k=8$ configurations retrieve every record. The $k=6$ pair has different retrieved length sequences on $2$ questions.

Learned EM gains are $8.0$--$23.6$ percentage points for six configurations at $\bar r\approx0.83$--$0.91$, versus $-0.2/-0.4$ near $0.95$ and $-2.2$ to $+3.6$ at $0.68$--$0.77$. Mean learned-minus-rule position gaps are $0.05\%$--$1.64\%$ of the hard reference's body positions (per-question maximum $5.95\%$). The main run's equal-position subset ($236$ questions, $118$ owners) retains an EM gain of $8.05$ percentage points, paired $95\%$ interval $[3.81,12.29]$. This deployment-selected subset preserves the original maintenance timing (Appendix~\ref{app:D}).

\begin{table*}[!t]
\centering
\begin{minipage}[t]{\dimexpr\textwidth/2-\columnsep/2\relax}
\vspace{0pt}
\centering
\captionsetup{type=table}
\caption{Training-component ablations (MSC-derived 535, frozen Qwen2.5-7B, $k=8$). Trained variants use one seed and the final checkpoint. Updates are actual counts. F1/EM (percent) and $R_{\mathrm{all}}$ are owner-weighted. Interventions and NLL appear in Table~\ref{tab:E5}.}
\label{tab:5}
\footnotesize
\setlength{\tabcolsep}{3pt}
\begin{tabularx}{\linewidth}{@{}>{\raggedright\arraybackslash}Xrrrr@{}}
\toprule
Configuration & Updates & $R_{\mathrm{all}}$ & F1 & EM \\
\midrule
\textbf{Full HasMem} & 5465 & $0.936$ & 95.3 & 84.3 \\
No forced softening or target band & 6068 & $0.752$ & 82.4 & 27.1 \\
No forced softening & 5703 & $0.760$ & 83.8 & 30.6 \\
No target band & 5319 & $0.819$ & 91.7 & 47.9 \\
Frozen Writer without hinge & 6868 & $0.811$ & 87.0 & 34.9 \\
No Writer hinge & 6724 & $1.000$ & 90.9 & 88.1 \\
No KEEP-wrong weighting & 5572 & $0.914$ & 96.1 & 81.7 \\
No Reader & 5085 & $0.917$ & 96.4 & 82.1 \\
No Global readout & 5181 & $0.890$ & 93.9 & 75.6 \\
Hard reference & No training & $1.000$ & 90.9 & 88.1 \\
\bottomrule
\end{tabularx}
\end{minipage}%
\hfill
\begin{minipage}[t]{\dimexpr\textwidth/2-\columnsep/2\relax}
\vspace{0pt}
\centering
\captionsetup{type=table}
\caption{Readout gains on MSC-derived 535. $\Delta$EM is full readout minus the indicated disabled pathway (question-weighted percentage points). $R_{\mathrm{all}}$ is owner-weighted. $\rho$ is the rule target width ratio. Owner-level exact $p$ values use separate Bonferroni families of two learned and eight rule comparisons (bold indicates adjusted $p<0.05$). Full results appear in Tables~\ref{tab:G2}--\ref{tab:G3}.}
\label{tab:6}
\footnotesize
\setlength{\tabcolsep}{2pt}
\begin{tabular}{lccccc}
\toprule
& & \multicolumn{2}{c}{Reader} & \multicolumn{2}{c}{Global} \\
\cmidrule(lr){3-4}\cmidrule(lr){5-6}
Allocation & $R_{\mathrm{all}}$ & $\Delta$EM & $p$ & $\Delta$EM & $p$ \\
\midrule
Learned & $0.936$ & $+0.37$ & 1.0000 & $+1.12$ & 0.1406 \\
\midrule
Rule-based $\rho=0.91$ & $0.909$ & $+1.87$ & 0.1702 & $+1.87$ & 0.2471 \\
Rule-based $\rho=0.85$ & $0.826$ & $+1.12$ & 1.0000 & $+1.12$ & 1.0000 \\
Rule-based $\rho=0.75$ & $0.775$ & $+1.87$ & 0.0508 & $+2.24$ & \textbf{0.0146} \\
Rule-based $\rho=0.60$ & $0.752$ & $+0.00$ & 1.0000 & $+0.00$ & 1.0000 \\
\bottomrule
\end{tabular}
\end{minipage}
\end{table*}

\subsubsection{Ablation and Mechanism Analysis}
\label{sec:ablations}
\paragraph{Training components.}
Table~\ref{tab:5} ablates the curriculum, Writer training, and readout components. Variants without forced softening or the target band have lower F1, EM, and realized budgets. Removing Writer hinge yields all KEEP and matches the hard reference. The main MSC-derived trajectory contains $1470$ KEEP, $516$ SHRINK, and no EXPAND actions.

\FloatBarrier
\paragraph{Readout pathways.}
Table~\ref{tab:6} holds model parameters and per-question positions fixed within each allocation condition, disabling Reader or Global increments during answering. Learned allocation preserves Controller actions and Bank/Global trajectories. Rule-based allocation uses four target widths. Global updates and Writer conditioning remain active.

Global at rule target $\rho=0.75$ gains $2.24$ percentage points in EM and passes correction ($p=0.0146$). Reader gains $1.87$ percentage points ($p=0.0508$). All other comparisons have adjusted $p\ge0.05$. Disabling either pathway increases NLL throughout. EM gains depend on allocation and deployment.

\FloatBarrier
\section{Limitations}
\label{sec:limitations}
MSC-derived uses development-split sentence reconstruction with at most $7$ records per owner and a rule-based ceiling of $100.0$. Longer histories and independent testing are needed to assess accumulation, interference, scaling, and generalization. Higher F1 accompanies fewer abstentions, lower EM, and higher NLL than the hard reference.

In Table~\ref{tab:4}, discrete position residuals and maintenance timing limit attribution to width allocation. Cross-training comparisons additionally vary in seeds, updates, and checkpoints. Storage, GPU memory, and latency depend on vector precision, module parameters, caches, and implementation overhead as well as position counts. Table~\ref{tab:E4} evaluates Gaussian-input-trained and rule-trained models under hard-input, Controller-driven deployment. Gaussian-input and rule-action deployment remain untested.

Fixed-model readout interventions estimate answering-stage pathway effects. Retrained module effects across seeds, configuration--module interactions, and EXPAND's independent contribution require controlled evaluation. Owner-clustered intervals capture evaluation-sample variability and assume independent owners representative of the target population.

The local LongMemEval-S judge requires validation against the reference evaluation configuration. Different realized budgets and integration protocols limit equal-resource comparisons with external systems.

\section{Conclusion}
\label{sec:conclusion}
HasMem learns per-record width adjustment, re-encoding, and readout around a frozen LLM, starting from verifiable token-aligned embeddings. All-KEEP trajectories preserve hard-prompt equivalence under identical retrieval and input construction. At approximately matched per-question target body budgets, six configurations at mean per-record retention $0.83$--$0.91$ exceed rule-based re-encoding by $8.0$--$23.6$ EM percentage points. The main configuration improves MSC-derived F1 by $4.4$ percentage points at $93.6\%$ of the hard reference's framed positions. LongMemEval-S F1 improves by $5.6$ percentage points and NLL decreases by $6.983$. These gains coexist with lower EM. Width-allocation and readout benefits depend on budget and deployment conditions, supporting further study of continual softening and adapted readout.

\section*{Generative Artificial Intelligence (AI) Usage Disclosure}
The authors determined the research question, core framework, and methodological design. Generative AI tools assisted with research ideation, parts of experimental refinement, code implementation and experiment execution, parts of result analysis and checks of formal statements, selected literature searches and preliminary categorization, and drafting portions of the paper, translation, and language editing. The authors screened, verified, and integrated the literature; reviewed and revised the plans, code, analyses, and manuscript; and take responsibility for the final research judgments, citations, and content.

\bibliographystyle{abbrvnat}
{\small\bibliography{library}}
\clearpage
\appendix
\setlength{\textfloatsep}{8pt plus 2pt minus 2pt}
\setlength{\dbltextfloatsep}{8pt plus 2pt minus 2pt}
\setlength{\floatsep}{8pt plus 2pt minus 2pt}
\setlength{\dblfloatsep}{8pt plus 2pt minus 2pt}
\setlength{\intextsep}{8pt plus 2pt minus 2pt}
\renewcommand{\thetable}{\Alph{section}\arabic{table}}
\renewcommand{\theHtable}{\Alph{section}.\arabic{table}}
\setcounter{table}{0}
\section{Supplementary Diagnostics of External Memory Systems}
\label{app:A}

\subsection{Answering and Paraphrasing Sensitivity on MSC-derived}
\label{app:A1}

\begin{table*}[!tbp]
\centering
\caption{Diagnostic results of external-system adaptations on MSC-derived 535 with frozen Qwen2.5-7B and full-sentence answers.}
\label{tab:A1}
\footnotesize
\setlength{\tabcolsep}{3pt}
\renewcommand{\arraystretch}{1.12}
\begin{tabularx}{\textwidth}{@{}XrrX@{}}
\toprule
Configuration & MSC F1 & MSC EM & Writing and readout \\
\midrule
Verbatim window & 93.04 & 86.19 & Verbatim text \\
BM25 retrieval ($k=8$) & 87.91 & 78.73 & Verbatim text \\
Hard embedding retrieval & 87.86 & 77.99 & Verbatim text \\
A-MEM adaptation & 85.57 & 75.75 & One write per event \\
Mem0 adaptation & 5.65 & 0.00 & One write per event \\
LLMLingua-2 (official, single pass) & 19.67 & 0.00 & One token-pruning pass before readout at nominal retention $0.33$ \\
\bottomrule
\end{tabularx}
\end{table*}

This appendix reports the reconstruction performance and integration protocols of external-system adaptations on MSC-derived 535. The probe is sensitive to the preservation of identifiers and original wording. Its scores reflect memory construction, retrieval, and answer generation jointly.

Table~\ref{tab:A1} reports averages with equal weight for each of the $268$ owners. BM25 and hard embedding retrieval use $k=8$. Mem0 and A-MEM receive individual events but construct memories at different granularities. LLMLingua-2 compresses text before readout. Paired results for hard embedding retrieval and LLMLingua-2 with the same backbone appear in Table~\ref{tab:A3}.

MSC-derived uses the complete stored record value as the gold answer, whereas a short answer may return only one span. For example, \textquotedblleft I go to concerts every weekend.\textquotedblright{} and \textquotedblleft concerts\textquotedblright{} convey related information but differ under full-sentence lexical scoring.

In Table~\ref{tab:A1}, verbatim-text references achieve higher F1 and EM than the evaluated external memory adaptations. Semantic question answering and long-term memory management require corresponding task-specific evaluations.

Mem0~\citep{chhikara2025mem0} and LLMLingua-2~\citep{pan2024llmlingua} obtain lexical EM of $0$ on this field-based question-answering probe. Identifier preservation, paraphrasing, and integration choices may contribute to these scores. The available comparisons do not isolate their effects. The local Memory-R1~\citep{yan2026memory} adaptation lacks a configuration trained with reinforcement learning and is excluded from this table.

\subsection{Complete Supplementary Results on LongMemEval-S}
\label{app:A2}

\begin{table*}[!tbp]
\centering
\caption{Descriptive comparisons on all 500 LongMemEval-S questions with frozen Qwen2.5-7B.}
\label{tab:A2}
\footnotesize
\setlength{\tabcolsep}{3pt}
\renewcommand{\arraystretch}{1.12}
\begin{tabularx}{\textwidth}{@{}XrrrX@{}}
\toprule
Method & Cover & Brief & Lexical F1 & Protocol \\
\midrule
Hard reference, same retrieval width & 10.60 & 9.80 & 3.39 & Retrieval settings shared with HasMem \\
HasMem (main) & 14.20 & 5.60 & 8.94 & Learned softening with the configuration in Table~\ref{tab:2} \\
LLMLingua-2 (official, single pass) & 26.20 & 1.80 & 9.23 & Whole-context compression at nominal ratio $0.33$ \\
\bottomrule
\end{tabularx}
\end{table*}

Table~\ref{tab:A2} summarizes the available comparisons on all $500$ questions. The Mem0 and Memory-R1 runs share state across questions and do not satisfy the question-level isolation protocol. Complete results with independent state for each question are unavailable for these two adaptations. MSC-derived uses each owner as the memory-isolation unit, as specified in Table~\ref{tab:A1}.

All three rows use frozen Qwen2.5-7B, the same question set, greedy decoding with a $64$-token answer budget, and the same rule-based scoring. HasMem and its hard reference share the controlled retrieval settings. LLMLingua-2 uses whole-context compression, with input construction and realized memory budgets that differ from the other two methods.

LLMLingua-2 obtains F1 of $9.23$, compared with $8.94$ for HasMem. These are descriptive results under their respective deployment protocols. LLMLingua-2 has higher Cover and lower Brief. These metrics measure answer coverage and coverage subject to a length constraint, respectively. Sentence reconstruction on MSC-derived and external question answering on LongMemEval-S address different evaluation objectives.

\subsection{Sentence Reconstruction with \mbox{HasMem} and External Compressors}
\label{app:A3}

\begin{table*}[!tbp]
\centering
\caption{Sentence reconstruction by HasMem and external compressors on MSC-derived 535. Differences are relative to the hard reference within each group.}
\label{tab:A3}
\footnotesize
\setlength{\tabcolsep}{3pt}
\renewcommand{\arraystretch}{1.12}
\begin{tabularx}{\textwidth}{@{}>{\raggedright\arraybackslash}p{3.45cm}l>{\raggedright\arraybackslash}p{1.65cm}Xrrrr@{}}
\toprule
Method & Backbone & \shortstack{Memory\\form} & \shortstack{Position\\ratio} & MSC F1 & $\Delta$F1 & MSC EM & $\Delta$EM \\
\midrule
Hard reference (HasMem) & Qwen2.5-7B & Token embeddings & $1.00\times$ & 90.87 & \textemdash & 88.06 & \textemdash \\
HasMem (main) & Qwen2.5-7B & Continuous slots & $0.936\times$ & 95.29 & $+4.42$ & 84.33 & $-3.73$ \\
\midrule
Hard embedding retrieval & Qwen2.5-7B & Text tokens & $1.00\times$ & 87.86 & \textemdash & 77.99 & \textemdash \\
LLMLingua-2 (official, single pass) & Qwen2.5-7B & Text tokens & $0.33\times$ (nominal) & 19.67 & $-68.19$ & 0.00 & $-77.99$ \\
\midrule
Hard embedding retrieval & Mistral-7B & Text tokens & $1.00\times$ & 72.11 & \textemdash & 31.34 & \textemdash \\
ICAE v2, one bank per context & Mistral-7B & Continuous slots & $1.48\times$ & 28.45 & $-43.66$ & 10.26 & $-21.08$ \\
ICAE v2, one bank per record & Mistral-7B & Continuous slots & $7.13\times$ & 0.62 & $-71.48$ & 0.00 & $-31.34$ \\
\bottomrule
\end{tabularx}
\end{table*}

Table~\ref{tab:A3} compares HasMem with two released compressors, pairing each with a hard reference that uses the same backbone and covers the same original events. On the token side, LLMLingua-2 applies its official single-pass, whole-context compression procedure at the nominal ratio of $0.33$. On the continuous-vector side, ICAE~\citep{ge2023context} uses official v2 weights and fixed slot banks. The ICAE comparison uses Mistral-7B-Instruct-v0.2, the backbone corresponding to its released weights, together with a separate text reference for that backbone. These comparisons evaluate each complete compression pipeline on sentence reconstruction.

Within each integration group, the method and its reference share the answer instruction, $64$-token generation budget, and lexical scoring, with owner-weighted aggregation. The HasMem group uses the main configuration and input construction of Table~\ref{tab:1}. External compressors use their respective integration protocols. ICAE shares its reference's top-$k$ retrieval. LLMLingua-2 compresses the whole context in event-arrival order, whereas its reference orders events by question similarity. Both cover the same events with different input orders. Instructions and questions remain in text form.

The evaluation uses the same $268$ owners and $535$ questions as Table~\ref{tab:E3}, with their records processed using the Qwen tokenizer. The memory-position ratio divides each method's unrounded owner-weighted mean position count by that of its hard reference on the same questions. Ratios for HasMem and ICAE are measured. The LLMLingua-2 pipeline does not record the post-compression token count, so its official nominal ratio is reported and marked accordingly. The Qwen text reference and LLMLingua-2 F1 and EM scores also appear in Table~\ref{tab:A1}.

Both external compressors score below their paired hard references on verbatim reconstruction (Table~\ref{tab:A3}). LLMLingua-2 targets task-agnostic textual fidelity and produces no exact matches here. ICAE uses $1.48\times$ and $7.13\times$ the hard reference's positions. Its scores reflect fixed slot counts, bank concatenation, prompt formats, and training-distribution adaptation jointly.

Retrieval chunks contain $128$ tokens, and each memory manager constructs entries according to its own protocol. Each table specifies the complete question set and state-isolation boundary used for evaluation.

\setcounter{table}{0}
\section{Target-Band Penalty Sweep}
\label{app:B}

\begin{table*}[!tbp]
\centering
\caption{Target-band penalty sweep on MSC-derived 535, with a target of $5465$ training updates per configuration.}
\label{tab:B1}
\footnotesize
\setlength{\tabcolsep}{3pt}
\renewcommand{\arraystretch}{1.12}
\begin{tabularx}{\textwidth}{@{}lrrrrrXl@{}}
\toprule
\shortstack{Backbone\\($k=8$)} & $\lambda_{\mathrm{band}}$ & F1 & EM & NLL & $R_{\mathrm{all}}$ & \shortstack{Hard reference\\F1 / EM / NLL} & $\Delta$F1 / $\Delta$EM \\
\midrule
0.5B & $0.0$ & 69.8 & 12.3 & 1.065 & $0.670$ & 59.4 / 51.1 / 0.187 & $+10.4$ / $-38.8$ \\
0.5B & $0.6$ & 58.9 & 7.1 & 1.598 & $0.604$ & 59.4 / 51.1 / 0.187 & $-0.4$ / $-44.0$ \\
0.5B & $1.5$ & 62.5 & 9.9 & 1.349 & $0.632$ & 59.4 / 51.1 / 0.187 & $+3.1$ / $-41.2$ \\
0.5B & $3.0$ & 61.1 & 9.9 & 1.417 & $0.637$ & 59.4 / 51.1 / 0.187 & $+1.7$ / $-41.2$ \\
0.5B & $6.0$ & 91.5 & 63.2 & 0.197 & $0.924$ & 59.4 / 51.1 / 0.187 & $+32.1$ / $+12.1$ \\
\midrule
1.5B & $0.0$ & 63.0 & 9.0 & 1.362 & $0.602$ & 90.4 / 80.8 / 0.039 & $-27.4$ / $-71.8$ \\
1.5B & $0.6$ & 74.9 & 17.0 & 0.924 & $0.665$ & 90.4 / 80.8 / 0.039 & $-15.5$ / $-63.8$ \\
1.5B & $1.5$ & 82.3 & 28.2 & 0.675 & $0.714$ & 90.4 / 80.8 / 0.039 & $-8.1$ / $-52.6$ \\
1.5B & $3.0$ & 86.5 & 49.4 & 0.241 & $0.909$ & 90.4 / 80.8 / 0.039 & $-4.0$ / $-31.3$ \\
1.5B & $6.0$ & 93.4 & 79.9 & 0.089 & $0.957$ & 90.4 / 80.8 / 0.039 & $+2.9$ / $-0.9$ \\
\midrule
3B & $0.0$ & 94.3 & 77.2 & 0.172 & $0.889$ & 84.7 / 74.6 / 0.167 & $+9.5$ / $+2.6$ \\
3B & $0.6$ & 92.2 & 72.0 & 0.247 & $0.892$ & 84.7 / 74.6 / 0.167 & $+7.5$ / $-2.6$ \\
3B & $1.5$ & 94.5 & 77.8 & 0.144 & $0.915$ & 84.7 / 74.6 / 0.167 & $+9.8$ / $+3.2$ \\
3B & $3.0$ & 94.2 & 57.8 & 0.186 & $0.909$ & 84.7 / 74.6 / 0.167 & $+9.4$ / $-16.8$ \\
3B & $6.0$ & 91.4 & 80.6 & 0.127 & $0.973$ & 84.7 / 74.6 / 0.167 & $+6.7$ / $+6.0$ \\
\midrule
7B & $0.0$ & 93.4 & 75.2 & 0.174 & $0.887$ & 90.9 / 88.1 / 0.074 & $+2.5$ / $-12.9$ \\
7B & $0.6$ & 87.5 & 38.4 & 0.467 & $0.788$ & 90.9 / 88.1 / 0.074 & $-3.4$ / $-49.6$ \\
7B & $1.5$ & 95.5 & 80.0 & 0.115 & $0.917$ & 90.9 / 88.1 / 0.074 & $+4.7$ / $-8.0$ \\
7B & $3.0$ & 92.6 & 90.5 & 0.063 & $0.992$ & 90.9 / 88.1 / 0.074 & $+1.7$ / $+2.4$ \\
7B & $6.0$ & 94.6 & 86.0 & 0.094 & $0.970$ & 90.9 / 88.1 / 0.074 & $+3.7$ / $-2.1$ \\
\midrule
14B & $0.0$ & 91.7 & 45.1 & 0.333 & $0.814$ & 97.4 / 93.3 / 0.045 & $-5.7$ / $-48.1$ \\
14B & $0.6$ & 91.9 & 49.8 & 0.305 & $0.814$ & 97.4 / 93.3 / 0.045 & $-5.5$ / $-43.5$ \\
14B & $1.5$ & 97.1 & 84.3 & 0.098 & $0.916$ & 97.4 / 93.3 / 0.045 & $-0.3$ / $-9.0$ \\
14B & $3.0$ & 97.4 & 93.3 & 0.045 & $1.000$ & 97.4 / 93.3 / 0.045 & $0.0$ / $0.0$ \\
14B & $6.0$ & 98.0 & 91.8 & 0.043 & $0.974$ & 97.4 / 93.3 / 0.045 & $+0.6$ / $-1.5$ \\
\bottomrule
\end{tabularx}
\end{table*}

Table~\ref{tab:B1} sweeps $\lambda_{\mathrm{band}}\in\{0,0.6,1.5,3,6\}$ across five backbones, using all $535$ MSC-derived questions and a target of $5465$ training updates. The target-band half-width is $\beta=0.12$, the forced-softening sampling probability decreases from $75\%$ to $15\%$, and the default penalty strength is $0.6$. Other training, checkpoint-selection, and evaluation settings are shared.

The relative position ratio $R_{\mathrm{all}}$ is the number of memory positions, including role framing, divided by that of the hard reference under the same configuration. The hard references use $101.1$ vectors for every backbone. Differences are relative to the hard reference with the same backbone and retrieval width. Two of the twenty-five configurations finish below the target update count, completing $5132$ updates for 1.5B with $\lambda_{\mathrm{band}}=0$ and $4386$ for 14B with $\lambda_{\mathrm{band}}=0.6$. The 14B configuration with $\lambda_{\mathrm{band}}=3.0$ has relative budget $1.000$. The Controller selects KEEP throughout, the causal gate remains closed, and all four metrics match the hard reference exactly.

Each configuration uses one seed. The eight-seed results in Table~\ref{tab:C1} provide a descriptive reference for 7B (F1 $90.9$--$96.1$, EM $77.4$--$88.1$). Corresponding cross-seed ranges are unavailable for the other backbones.

The relationship between penalty strength and EM varies by backbone. EM increases monotonically for 1.5B, peaks at $\lambda_{\mathrm{band}}=3.0$ and then decreases for 14B, improves at the largest strength for 0.5B, and is nonmonotonic for 3B and 7B.

The default-strength 14B configuration in Table~\ref{tab:B1} completes $4386$ updates, while the independent training run used in Table~\ref{tab:2} completes $3752$. In this sweep, default-strength EM is $43.5$ percentage points below the hard reference. At $\lambda_{\mathrm{band}}=1.5$, F1 approaches the hard reference at a relative budget of $0.916$, but EM remains $9.0$ points lower and answer NLL is $0.098$, compared with $0.045$ for the hard reference.

Realized budget and EM vary jointly. The seven configurations with budgets below $0.72$ have EM of $7.1$--$28.2$, the ten configurations with budgets between $0.81$ and $0.92$ have EM of $45.1$--$84.3$, and the six configurations with budgets above $0.95$ have EM of $79.9$--$93.3$. The highest-EM configurations for the five backbones have relative budgets of $0.924$, $0.957$, $0.973$, $0.992$, and $1.000$, respectively. Performance can also vary at similar budgets. The five 3B configurations span budgets of $0.889$--$0.973$ and an EM range of $22.8$ points.

Relative to their respective hard references, both F1 and EM improve for 0.5B at $\lambda_{\mathrm{band}}=6.0$ and for 3B at $\lambda_{\mathrm{band}}=0,1.5,6.0$. The 3B configurations at $\lambda_{\mathrm{band}}=1.5,6.0$ also have lower NLL ($0.144$ and $0.127$) than the hard reference ($0.167$). NLL is slightly higher than the hard reference for 0.5B at $\lambda_{\mathrm{band}}=6.0$ and 3B at $\lambda_{\mathrm{band}}=0$. For 7B, $\lambda_{\mathrm{band}}=3.0$ yields EM of $90.5$ at relative budget $0.992$, compared with $88.1$ for the hard reference.

\setcounter{table}{0}
\section{Cross-Seed Variation in Training and Checkpoint Selection}
\label{app:C}

\begin{table}[!htbp]
\centering
\caption{Eight-seed evaluation on MSC-derived 535 (7B, $k=8$). Each run completes $5465$ updates, including $5081$ policy updates. Checkpoints are selected by in-band training loss.}
\label{tab:C1}
\footnotesize
\setlength{\tabcolsep}{3pt}
\renewcommand{\arraystretch}{1.12}
\begin{tabularx}{\columnwidth}{@{}Xrrrrr@{}}
\toprule
Seed & \shortstack{Selected\\update} & $R_{\mathrm{all}}$ & F1 & EM & Answer NLL \\
\midrule
$2026091331$ & 385 & $1.000$ & 90.9 & 88.1 & 0.074 \\
$2026091401$ & 411 & $0.964$ & 96.1 & 82.3 & 0.110 \\
$2026091701$ & 4906 & $0.916$ & 93.5 & 77.4 & 0.161 \\
$2026091702$ & 3801 & $0.918$ & 93.6 & 79.3 & 0.134 \\
$2026091703$ & 417 & $0.968$ & 93.7 & 79.7 & 0.095 \\
$2026091711$ & 422 & $0.966$ & 96.0 & 86.6 & 0.072 \\
$2026091712$ & 4922 & $0.935$ & 94.6 & 82.8 & 0.124 \\
$2026091713$ & 426 & $0.966$ & 95.5 & 84.5 & 0.094 \\
\midrule
Mean & \textemdash & $0.954$ & 94.2 & 82.6 & 0.108 \\
\bottomrule
\end{tabularx}
\end{table}

Eight independent retraining runs use fixed random seeds, a curriculum indexed by update count, the same training budget, and the same checkpoint-selection rule based on in-band training loss. Each run completes $5465$ updates and is evaluated at its selected checkpoint. Variation in the selected update and realized position budget reflects this training and selection procedure jointly.

The hard reference at the same retrieval width has F1/EM/NLL of $90.9/88.1/0.074$. Selected update counts include $384$ warm-up updates. The relative budget is the total number of memory positions, including role framing, relative to the hard reference under the same configuration.

For seed $2026091331$, the in-band rule selects update $385$, whose trajectory is all KEEP and whose results match the hard reference. The sample standard deviations of F1, EM, answer NLL, and relative budget are $1.73$ percentage points, $3.70$ percentage points, $0.031$, and $0.029$, respectively.

\setcounter{table}{0}
\section{Paired Sensitivity Analysis of Budget Residuals}
\label{app:D}

\begin{table*}[!tbp]
\centering
\caption{Paired EM differences and equal-body-position subsets for all configurations in Table~\ref{tab:4}. Differences are learned minus rule-based EM, in percentage points.}
\label{tab:D1}
\footnotesize
\setlength{\tabcolsep}{3pt}
\renewcommand{\arraystretch}{1.12}
\begin{tabularx}{\textwidth}{@{}Xlll@{}}
\toprule
Configuration & \shortstack{All questions\\$\Delta$EM [95\% interval]} & \shortstack{Subset\\questions / owners} & \shortstack{Equal-position subset\\$\Delta$EM [95\% interval]} \\
\midrule
Seed 2026091703 & -0.19 [-1.12, 0.75] & 511 / 256 & +0.00 [-0.59, 0.59] \\
Seed 2026091401 & -0.37 [-1.12, 0.37] & 512 / 256 & -0.20 [-0.59, 0.00] \\
Original configuration & +17.94 [14.55, 21.39] & 236 / 118 & +8.05 [3.81, 12.29] \\
Seed 2026091702 & +13.83 [10.09, 17.54] & 248 / 124 & +6.45 [1.61, 11.29] \\
Seed 2026091701 & +23.55 [19.66, 27.43] & 253 / 127 & +17.79 [12.20, 23.62] \\
\texttt{full} (fixed update count) & +19.44 [15.54, 23.26] & 177 / 89 & +12.43 [5.17, 19.66] \\
\texttt{full}, $k=6$ & +9.16 [6.16, 12.31] & 350 / 176 & +4.29 [1.15, 7.45] \\
Gaussian-input training & +8.04 [5.23, 10.86] & 366 / 183 & +5.74 [3.01, 8.47] \\
\texttt{full}, width adjustment $20\%$ & -1.50 [-5.41, 2.43] & 221 / 111 & +0.90 [-5.00, 6.82] \\
\texttt{full} $-$ target band & +3.55 [0.56, 6.54] & 291 / 146 & +1.37 [-2.74, 5.17] \\
\texttt{full}, final target $30\%$ & +2.99 [0.75, 5.23] & 447 / 224 & +1.12 [-0.67, 3.12] \\
\texttt{full} $-$ forced softening & -2.24 [-5.23, 0.75] & 247 / 124 & -0.40 [-4.47, 4.03] \\
\bottomrule
\end{tabularx}
\end{table*}

For the original configuration in Table~\ref{tab:4}, the rule-based trajectory uses fewer body positions on $299$ questions. Across all $535$ questions, it uses $0.94$ fewer positions on average and at most $5$ fewer positions on any question.

Table~\ref{tab:D1} uses question-weighted differences and the owner-clustered paired bootstrap from Section~\ref{sec:setup}. Every full-set comparison contains $535$ questions from $268$ owners. The equal-position subset retains questions for which the two trajectories use exactly the same number of body positions, preserving their original predictions and maintenance trajectories. Intervals are descriptive, pointwise 95\% intervals. Subsets are selected from deployment outcomes and therefore change the estimand.

Differences on equal-position subsets describe the two existing trajectories under that condition. Online maintenance and re-encoding after the complete history still differ in timing. Isolating the effect of the width-allocation rule requires joint control of the maintenance trajectory and realized budget.

\setcounter{table}{0}
\section{Training Settings and Supplementary Experiments}
\label{app:E}

\subsection{Training Settings for Backbone and Retrieval-Width Configurations}
\label{app:E1}

\begin{table}[!htbp]
\centering
\caption{Completed training updates for the backbone and retrieval-width configurations in Tables~\ref{tab:1} and~\ref{tab:2}.}
\label{tab:E1}
\footnotesize
\setlength{\tabcolsep}{3pt}
\renewcommand{\arraystretch}{1.12}
\begin{tabularx}{\columnwidth}{@{}Xrr@{}}
\toprule
Backbone & $k$ & Training updates \\
\midrule
Qwen2.5-0.5B & 8 & 5683 \\
Qwen2.5-1.5B & 4 & 2907 \\
Qwen2.5-1.5B & 8 & 4949 \\
Qwen2.5-3B & 8 & 5065 \\
Qwen2.5-7B & 2 & 5879 \\
Qwen2.5-7B & 4 & 5825 \\
Qwen2.5-7B & 6 & 5476 \\
Qwen2.5-7B & 8 & 5465 \\
Qwen2.5-7B & 12 & 5295 \\
Qwen2.5-14B & 8 & 3752 \\
Mistral-7B & 8 & 5046 \\
\bottomrule
\end{tabularx}
\end{table}

The configurations in Table~\ref{tab:E1} use the MSC-derived training set, the same wall-clock budget, and the main seed $2026091331$. Table~\ref{tab:2} reports all configurations on LongMemEval-S 500. Table~\ref{tab:1} reports the Qwen2.5-7B configurations with $k=2,6,8,12$ on MSC-derived 535.

\subsection{Seed and Hyperparameter Configurations}
\label{app:E2}

\begin{table*}[!tbp]
\centering
\caption{Seed and hyperparameter configurations on MSC-derived 535 (Qwen2.5-7B, $k=8$, unless specified). Updates are listed as evaluated checkpoint / completed training.}
\label{tab:E2}
\footnotesize
\setlength{\tabcolsep}{3pt}
\renewcommand{\arraystretch}{1.12}
\begin{tabularx}{\textwidth}{@{}Xllrrrr@{}}
\toprule
Configuration & Seed & \shortstack{Updates\\evaluated / completed} & F1 & EM & NLL & $R_{\mathrm{all}}$ \\
\midrule
\texttt{full} + always-on readout gate & 2026091331 & 5465 / 5465 & 95.1 & 80.0 & 0.135 & $0.914$ \\
Additional hard anchoring & 2026091401 & 426 / 5465 & 94.3 & 81.0 & 0.100 & $0.966$ \\
Additional hard anchoring & 2026091701 & 428 / 5465 & 92.5 & 78.2 & 0.114 & $0.959$ \\
Additional hard anchoring & 2026091702 & 422 / 5465 & 97.0 & 89.7 & 0.059 & $0.967$ \\
Additional hard anchoring & 2026091703 & 433 / 5465 & 93.7 & 82.6 & 0.091 & $0.968$ \\
Final target frequency $10\%$ & 2026091701 & 4896 / 5465 & 92.7 & 84.0 & 0.096 & $0.962$ \\
Final target frequency $10\%$ & 2026091702 & 4287 / 5465 & 91.1 & 80.4 & 0.174 & $0.948$ \\
Final target frequency $20\%$ & 2026091331 & 5121 / 5121 & 95.8 & 81.2 & 0.122 & $0.910$ \\
Final target frequency $30\%$ & 2026091331 & 5363 / 5363 & 83.7 & 34.5 & 0.638 & $0.799$ \\
Stronger band, 3B & 2026091331 & 5317 / 5317 & 90.8 & 80.6 & 0.144 & $0.976$ \\
Stronger band, 14B & 2026091331 & 4107 / 4107 & 96.9 & 69.8 & 0.108 & $0.909$ \\
Stronger band, 7B $k=4$ & 2026091331 & 5979 / 5979 & 91.7 & 86.2 & 0.135 & $0.971$ \\
Llama-3.1-8B, default band & 2026091331 & 5024 / 5024 & 91.1 & 47.6 & 0.347 & $0.789$ \\
Stronger band, Llama-3.1-8B & 2026091331 & 5782 / 5782 & 98.6 & 86.4 & 0.044 & $0.948$ \\
Final target frequency $0.32$ & 2026091401 & 5461 / 5465 & 94.0 & 88.6 & 0.078 & $0.972$ \\
Final target frequency $0.32$ & 2026091701 & 5283 / 5465 & 93.3 & 84.7 & 0.121 & $0.949$ \\
Final target frequency $0.32$ & 2026091702 & 5357 / 5465 & 97.0 & 90.5 & 0.062 & $0.968$ \\
Final target frequency $0.32$ & 2026091703 & 5328 / 5465 & 95.0 & 89.7 & 0.076 & $0.976$ \\
Width adjustment $20\%$ & 2026091331 & 4809 / 4809 & 84.7 & 34.5 & 0.513 & $0.688$ \\
Final target frequency $10\%$ & 2026091331 & 4921 / 4921 & 95.3 & 80.0 & 0.148 & $0.899$ \\
Default settings (last update) & 2026091401 & 5274 / 5274 & 93.3 & 78.5 & 0.217 & $0.908$ \\
Temporal weak-teacher selection & 2026091331 & 5040 / 5040 & 95.4 & 78.5 & 0.161 & $0.916$ \\
Final forced-sampling probability $5\%$ & 2026091331 & 5887 / 5887 & 96.9 & 86.2 & 0.084 & $0.948$ \\
\bottomrule
\end{tabularx}
\end{table*}

Additional hard anchoring adds a loss with weight $0.5$ whenever the original Writer hinge is computed. For each training question, it takes the positive part of the SHRINK-path gold-answer NLL minus the hard-path NLL and averages across questions. Gradients do not propagate through the hard path. The stronger-band setting uses $\lambda_{\mathrm{band}}=3.0$, $\beta=0.08$, and $(f_0,f_1)=(0.40,0.05)$, with the target frequency unchanged at $0.50\to0.22$. The final-target-frequency $0.32$ group also uses $\lambda_{\mathrm{band}}=3.0$ and $\beta=0.08$, while retaining the default forced-sampling probabilities of $0.75\to0.15$. These targets constrain the exponential moving average (EMA) of the predicted non-KEEP frequency. Realized position retention is determined by the trajectory. The temporal weak teacher prioritizes candidates that first execute SHRINK earlier, breaking ties by answer degradation and then position savings.

Update counts in Table~\ref{tab:E2} include $384$ warm-up updates. The additional-hard-anchoring group, the final-target-frequency $0.32$ group, and the two final-$10\%$ runs that complete $5465$ updates select the checkpoint with lowest training loss among those with non-KEEP prediction EMA in $[0.22,0.40]$. Other rows use the final update. The eight-seed results under default settings appear in Table~\ref{tab:C1}. All metrics are owner-weighted, and relative budget denotes $R_{\mathrm{all}}$.

Each configuration is compared with the hard reference using the same backbone and $k$. The hard reference's F1/EM/NLL is $90.9/88.1/0.074$ for the default 7B configuration, $84.7/74.6/0.167$ for 3B, $97.4/93.3/0.045$ for 14B, $91.1/88.4/0.096$ for 7B with $k=4$, and $95.8/92.9/0.032$ for Llama-3.1-8B. The default-settings row for seed $2026091401$ is an independent run completing $5274$ updates. The run with the same seed in Table~\ref{tab:C1} completes $5465$ and selects update $411$. The fixed-update, final-$10\%$ group contains two independent runs.

The four-seed mean EM of the additional-hard-anchoring group is $82.9$, below the hard reference's $88.1$. The final-target-frequency $0.32$ group has mean EM $88.4$. Three of its four runs match or exceed the hard reference, at relative budgets of $0.949$--$0.976$. Under the stronger band, 3B improves both F1 and EM relative to its hard reference. Llama-3.1-8B and 7B with $k=4$ improve F1 but have lower EM, and 14B has lower scores on both metrics. The stronger band changes penalty strength, band half-width, and the forced-sampling schedule jointly. These results reflect training settings, completed updates, and realized position budgets.

\subsection{Action-Policy Comparisons}
\label{app:E3}

\begin{table}[!htbp]
\centering
\caption{Action-policy comparisons on MSC-derived 535 with frozen Qwen2.5-7B and $k=8$.}
\label{tab:E3}
\footnotesize
\setlength{\tabcolsep}{3pt}
\renewcommand{\arraystretch}{1.12}
\begin{tabularx}{\columnwidth}{@{}XXrrr@{}}
\toprule
Method & Action rule & $R_{\mathrm{all}}$ & F1 & EM \\
\midrule
HasMem (original main run) & Controller-selected actions & $0.936$ & 95.3 & 84.3 \\
Hard reference & Original token embeddings & $1.000$ & 90.9 & 88.1 \\
Fixed & SHRINK at every step & $0.75$ & 80.0 & 22.9 \\
Random & Uniform sampling over three actions & $0.98$ & 90.6 & 50.4 \\
\bottomrule
\end{tabularx}
\end{table}

The comparisons in Table~\ref{tab:E3} share the answer instruction and scoring protocol. F1 and EM are owner-weighted. Relative budget is the ratio $R_{\mathrm{all}}$ of all memory positions, including role framing, to the hard reference. Table~\ref{tab:4} compares allocation strategies under approximately matched per-question target body budgets.

The random-action condition shares the trained parameters and Writer and samples uniformly from $\{\mathrm{KEEP},\mathrm{SHRINK},\mathrm{EXPAND}\}$. HasMem has EM gains of $34.0$ and $61.4$ percentage points over random and fixed actions, respectively. These differences jointly reflect action decisions, final widths, and state trajectories. The all-KEEP \texttt{noshrink} path shares the hard reference's implementation branch, so its results follow from their equivalence.

\subsection{Common-Deployment Evaluation of Training Recipes}
\label{app:E4}

\begin{table*}[!tbp]
\centering
\caption{Common-deployment evaluation of four training recipes with frozen Qwen2.5-7B and $k=8$.}
\label{tab:E4}
\footnotesize
\setlength{\tabcolsep}{3pt}
\renewcommand{\arraystretch}{1.12}
\begin{tabularx}{\textwidth}{@{}>{\raggedright\arraybackslash}p{2.6cm}Xrrrrrr@{}}
\toprule
Configuration & Training and deployment & \shortstack{MSC\\F1} & \shortstack{MSC\\EM} & \shortstack{MSC\\NLL} & \shortstack{Memory\\vectors} & \shortstack{LME-S\\F1} & \shortstack{LME-S\\NLL} \\
\midrule
Hard reference & Original token embeddings, without softening & 90.9 & 88.1 & 0.074 & 101.1 & 3.4 & 12.257 \\
\texttt{full} (fixed update count) & HasMem, $5465$ updates including $5081$ policy updates, final checkpoint & 94.0 & 74.8 & 0.172 & 89.8 & 9.8 & 5.260 \\
Gaussian-input training & Gaussian inputs during training and hard initialization at evaluation & 93.6 & 58.4 & 0.250 & 87.9 & 8.7 & 5.481 \\
Fixed-ratio rule training ($0.91$) & Rule-based training with action imitation and Controller at evaluation & 98.5 & 78.4 & 0.067 & 91.9 & 9.1 & 4.794 \\
Retention-based rule training ($0.91$) & Highest-retention entry selection during training and Controller at evaluation & 97.7 & 73.7 & 0.099 & 91.9 & 9.5 & 5.065 \\
\bottomrule
\end{tabularx}
\end{table*}

Table~\ref{tab:E4} evaluates four training recipes on MSC-derived 535 and LongMemEval-S 500. All four use the final checkpoint after $5465$ training updates, including $5081$ policy updates. Fixed-ratio and retention-based rule training additionally disable hinge, forced softening, and the target band, and use an action-imitation loss. Every model is evaluated from hard token embeddings with Controller-selected actions under the same deployment protocol.

The same trained models are evaluated on both complete question sets. Memory vectors denote the owner-weighted total number of positions, including role framing, on MSC-derived 535. LongMemEval-S columns report lexical F1 and gold-answer NLL, measuring generated-text overlap and the conditional likelihood of the gold answer, respectively.

All four softening configurations have lower MSC-derived EM, higher LongMemEval-S lexical F1, and lower LongMemEval-S answer NLL than the hard reference. The Gaussian-input-trained model experiences an input-distribution shift when deployed from hard embeddings. The two rule-trained models also use more memory positions than the full-curriculum model.

Under the same answer instruction, the fixed-ratio rule-trained model never outputs UNKNOWN on the $535$ MSC-derived questions. Its question-weighted F1/EM is $98.5/78.3$. Abstention counts and scores on answered subsets for the main configuration and the hard reference appear in Section~\ref{sec:hard-comparison}.

\subsection{Retraining Ablations}
\label{app:E5}

\begin{table*}[!tbp]
\centering
\caption{Training interventions and answer NLL for the ablations in Table~\ref{tab:5} (MSC-derived 535, frozen Qwen2.5-7B, $k=8$). NLL is owner-weighted. Configurations follow the order in Table~\ref{tab:5}, which reports training updates, $R_{\mathrm{all}}$, F1, and EM.}
\label{tab:E5}
\footnotesize
\setlength{\tabcolsep}{3pt}
\renewcommand{\arraystretch}{1.12}
\begin{tabularx}{\textwidth}{@{}>{\raggedright\arraybackslash}p{3.05cm}Xr@{}}
\toprule
Configuration & Training intervention & NLL \\
\midrule
\textbf{Full HasMem} & Full curriculum and all modules & 0.115 \\
No forced softening or target band & Remove forced softening and the target-band penalty & 0.621 \\
No forced softening & Remove forced softening & 0.571 \\
No target band & Remove the target-band penalty on non-KEEP prediction frequency & 0.285 \\
Frozen Writer without hinge & Freeze the warmed-up Writer and disable Writer hinge, retaining forward re-encoding & 0.488 \\
No Writer hinge & Disable Writer hinge, with an all-KEEP evaluation trajectory & 0.074 \\
No KEEP-wrong weighting & Remove KEEP-wrong weighting & 0.114 \\
No Reader & Remove low-rank readout adaptation & 0.089 \\
No Global readout & Disable Global readout during training, retaining state updates and Writer state input & 0.193 \\
Hard reference & Original token embeddings without softening & 0.074 \\
\bottomrule
\end{tabularx}
\end{table*}

MSC-derived ablation details appear in Table~\ref{tab:E5}.

\paragraph{LongMemEval-S retraining ablations.}
On all $500$ questions with frozen Qwen2.5-7B and $k=8$, disabling Soft Bank, disabling the Global readout increment, and removing Reader yield lexical F1 of $2.8/2.7$, $5.2/4.4$, and $7.3/7.2$, respectively, listed in seed order $2026091331/2026091401$. The full model trained with a fixed update count (Table~\ref{tab:E4}, seed $2026091331$) achieves $9.8$. Completed training updates differ across these groups. Their results jointly reflect module configuration, memory widths, and state trajectories.

\section{Training Objectives and Supervision}
\label{app:F}

This appendix specifies warm-up regression, policy-label search, curriculum penalties, and sample weighting. Hyperparameter values are reported with the experimental settings in Appendix~\ref{app:G}.

\paragraph{Warm-up.}
For each owner, intervention branches $a=0,1,2$ correspond to KEEP, SHRINK, and EXPAND. Their quality $Q_a$ and cumulative-position score $J_a$ are defined in Equations~\ref{eq:quality}--\ref{eq:branch-score}, using the owner's question set $Q$ and answer NLLs $\ell_{a,q}$ and $\ell_{h,q}$.
The representation and readout modules minimize mean $Q_a$ across the three branches. Controller regresses the relative targets $(J_1-J_0,J_2-J_0)$ with stopped gradients using a Smooth L1 loss. Writing $\operatorname{stopgrad}$ for an operator that preserves forward values and blocks gradients, the per-owner warm-up objective is
\begin{equation}
\label{eq:warmup}
\begin{aligned}
\mathcal L_{\mathrm{warm}}
&=\frac13\sum_{a=0}^{2}Q_a\\
&\quad+\operatorname{SmoothL1}\Bigl((\tilde\rho_1,\tilde\rho_2),\\
&\hspace{5em}\operatorname{stopgrad}(J_1-J_0,J_2-J_0)\Bigr).
\end{aligned}
\end{equation}

\paragraph{Policy supervision.}
After warm-up, we select action windows from training owners' histories, search candidate sequences with preset depth and beam width, and score them using each owner's training QA. A candidate sequence $b$ receives score $J_b$ from the same quality and cumulative-position objectives, with answer loss evaluated after the complete rollout. Candidates are ranked by $J_b+\lambda_{\mathrm{streak}}\min(s_b/s_{\mathrm{sat}},1)$, where $s_b$ is the KEEP streak at the window's end, $\lambda_{\mathrm{streak}}$ is its penalty weight, and $s_{\mathrm{sat}}>0$ is the saturation length. Let $\epsilon_{\mathrm{NLL}}\ge0$ be the per-question NLL tolerance. A qualifying branch must contain an actual SHRINK, keep each question's NLL no more than $\epsilon_{\mathrm{NLL}}$ above that of the KEEP branch sharing its prefix, strictly lower cumulative position cost, and strictly improve the search objective including the streak term. We select the lowest-scoring qualifying branch and use its first actual departure from KEEP as label $a^\star$. If no branch qualifies, the curriculum samples a weak label with probability $f(p)$ by selecting, among evaluated SHRINK branches that reduce cumulative position cost, the branch with the smallest maximum per-question damage and labeling its first SHRINK. KEEP is used when no weak label is obtained.

For supervised input $z=z_{i,t}$, action probabilities $\pi_\theta(a\mid z)$ follow Equation~\ref{eq:policy}.
For target $a^\star$ and sample weight $w$ defined below, the classification loss is
\begin{equation}
\label{eq:classification}
\mathcal L_\pi=-w\log\pi_\theta(a^\star\mid z).
\end{equation}

\paragraph{Forced-softening rate and target band.}
Policy progress $p$ is defined by elapsed wall-clock time. Let $T$ be the current time, $T_{\mathrm{start}}$ the policy-phase start time, and $T_{\mathrm{policy}}$ its available duration after reserving time for completion. Then
\begin{equation}
\label{eq:progress}
p=\operatorname{clip}\left(\frac{T-T_{\mathrm{start}}}{T_{\mathrm{policy}}},0,1\right).
\end{equation}
The operator $\operatorname{clip}(x,a,b)$ clips $x$ to $[a,b]$. The forced sampling probability for weak labels is $f(p)=f_0(1-p)+f_1p$, with initial and final probabilities $f_0$ and $f_1$. The target-band center $\tau(p)=\tau_0(1-p)+\tau_1p$ specifies the target frequency of non-KEEP predictions, with endpoints $\tau_0$ and $\tau_1$.

Let $n$ index owner samples in the policy phase, with supervised input $z_n$ and current prediction $\hat a_n=\arg\max_a\pi_\theta(a\mid z_n)$. We track the exponential moving average (EMA) $\bar u_n$ of non-KEEP prediction frequency, initialized at $\bar u_0=0$. For smoothing coefficient $\alpha_{\mathrm{EMA}}\in[0,1)$,
\begin{equation}
\label{eq:ema}
\bar u_n=\alpha_{\mathrm{EMA}}\bar u_{n-1}+(1-\alpha_{\mathrm{EMA}})\mathbf{1}[\hat a_n\ne\mathrm{KEEP}].
\end{equation}
The current sample uses $\bar u_{n-1}$ to compute the band penalty before updating the EMA. Let $\delta_n$ be the distance of $\bar u_{n-1}$ from $[\tau(p)-\beta,\tau(p)+\beta]$, where $\beta$ is the band half-width and $\lambda_{\mathrm{band}}$ the penalty weight. Then
\begin{equation}
\label{eq:band}
\begin{aligned}
\mathcal L_{\mathrm{band}}&=\lambda_{\mathrm{band}}\min(1,\delta_n/\beta)\\
&\quad\times\begin{cases}
\pi_\theta(\mathrm{KEEP}\mid z_n),&\bar u_{n-1}<\tau(p)-\beta,\\
1-\pi_\theta(\mathrm{KEEP}\mid z_n),&\bar u_{n-1}>\tau(p)+\beta,\\
0,&\text{otherwise}.
\end{cases}
\end{aligned}
\end{equation}
Forced labels supply supervision for rare non-KEEP actions, while the band penalty discourages deviations from the prescribed prediction frequency.

\paragraph{Quality constraints and sample weights.}
Let $\ell_{\mathrm{shrink},q}$ and $\ell_{\mathrm{keep},q}$ be answer NLLs for the selected SHRINK branch and its matched KEEP branch. With Writer quality-constraint weight $\lambda_{\mathrm{hinge}}$, the hinge loss is
\begin{equation}
\label{eq:hinge}
\begin{aligned}
\mathcal L_{\mathrm{hinge}}=\frac{\lambda_{\mathrm{hinge}}}{|Q|}\sum_{q\in Q}\Bigl[&\ell_{\mathrm{shrink},q}\\
&-\operatorname{stopgrad}(\ell_{\mathrm{keep},q})-\epsilon_{\mathrm{NLL}}\Bigr]_+.
\end{aligned}
\end{equation}
Both branches share the history preceding intervention. A SHRINK branch from the winning sequence is preferred. If none is available, we select the least-damaging evaluated candidate that lowers cumulative position cost and whose maximum per-question damage is at most $\epsilon_{\mathrm{cand}}$. The term is zero when no candidate is available. This constraint trains Writer through answer cross-entropy (CE).

Let $n^+$ and $n^-$ count qualifying and nonqualifying samples, including the current sample. $w_+^{\max}$ caps the base weight of qualifying positives, and $w_{\mathrm{weak}}(p)$ is the weak-label weight, which decreases linearly with progress. The base sample weight is
\begin{equation}
\label{eq:sample-weight}
w_0=\begin{cases}
\operatorname{clip}\bigl(n^-/\max(1,n^+),1,w_+^{\max}\bigr),&\text{qualifying},\\
w_{\mathrm{weak}}(p),&\text{weak label},\\
1,&\text{otherwise}.
\end{cases}
\end{equation}
If the hard reference fails to predict all gold-answer tokens under teacher forcing, including the answer-ending token used in training, or the visited entry's tag is absent from the targets of all training questions for that owner, we set $w=\min(w_{\max},\kappa_ww_0)$. Here $\kappa_w$ is a weight multiplier and $w_{\max}$ the final weight cap. Otherwise, $w=w_0$. Let $s$ be the number of consecutive KEEP actions preceding the supervised decision, corresponding to $s_t$ in Section~\ref{sec:controller}. Qualifying positives also receive the streak penalty
\begin{equation}
\label{eq:streak}
\begin{aligned}
\mathcal L_{\mathrm{streak}}={}&\lambda_{\mathrm{streak}}\mathbf{1}[\text{qualifying positive}]\\
&\times\min((s+1)/s_{\mathrm{sat}},1)\pi_\theta(\mathrm{KEEP}\mid z).
\end{aligned}
\end{equation}
The four terms form the policy objective in Equation~\ref{eq:policy-loss}.
Batch losses average equally over owners. Evaluation runs the trained Controller, Writer, Reader, and Global through forward state updates and width decisions.

\setcounter{table}{0}
\section{Experimental Settings and Statistical Details}
\label{app:G}

\subsection{Default Configuration}
\label{app:G1}

\begin{table*}[!tbp]
\centering
\caption{Default architecture, maintenance, and training settings. Progress $p$ follows the training schedule in Appendix~\ref{app:F}. Fixed-update variants are specified in Appendix~\ref{app:G2}.}
\label{tab:G1}
\footnotesize
\setlength{\tabcolsep}{4pt}
\renewcommand{\arraystretch}{1.06}
\begin{tabularx}{\textwidth}{@{}lXX@{}}
\toprule
Category & Setting & Default \\
\midrule
Architecture & Global/shared feature dimension $m$ and Controller MLP hidden dimension & $64$ and $64$ \\
& Reader and Global readout adaptation & Attention output projections in final $4$ layers, rank $r=4$ \\
& Added parameters and policy-phase trainable parameters & $4{,}402{,}627$ and $478{,}210$ \\
Maintenance & Width adjustment $\eta$ and minimum body width $K_{\min}$ & $0.1$ ($10\%$ before rounding) and $4$ \\
& KEEP-streak clipping $s_{\max}$ and round-robin interval $c_{\mathrm{rev}}$ & $8$ and $2$ \\
& Post-write maintenance steps & $12$ for Qwen2.5-0.5B/1.5B, $6$ for Qwen2.5-3B/7B/14B and Mistral-7B \\
& Global write-gate bias $b_r$ & $-4$, giving $\sigma(-4)\approx0.018$ \\
& Hard-equivalence tolerance for embeddings and logits & $10^{-4}$ \\
Optimization & Warmup and effective batch size & $384$ updates and $8$ owners \\
& Optimizer, learning rate, and weight decay & AdamW, $10^{-4}$, and $0$ \\
& Gradient-norm clipping and wall-clock budget & $1$ and approximately $6$ hours \\
& Time reserved before the job deadline & $20$ minutes \\
Supervision & Search depth and beam width & $2$ and $2$ \\
& Damage / position-cost weights $\lambda_{\mathrm{damage}}$, $\lambda_{\mathrm{cost}}$ & $4$ / $0.05$ \\
& Streak weight / saturation $\lambda_{\mathrm{streak}}$, $s_{\mathrm{sat}}$ & $0.02$ / $3$ \\
& NLL tolerance $\epsilon_{\mathrm{NLL}}$ & $10^{-4}$ \\
& Forced-sampling endpoints $(f_0,f_1)$ & $(0.75,0.15)$ \\
& Non-KEEP target endpoints $(\tau_0,\tau_1)$ & $(0.50,0.22)$ \\
& EMA coefficient $\alpha_{\mathrm{EMA}}$ & $0.95$ \\
& Band half-width / weight $\beta$, $\lambda_{\mathrm{band}}$ & $0.12$ / $0.6$ \\
& Writer hinge weight / candidate damage limit $\lambda_{\mathrm{hinge}}$, $\epsilon_{\mathrm{cand}}$ & $0.5$ / $0.2$ \\
& Positive base-weight cap $w_+^{\max}$ & $16$ \\
& Weak-label weight $w_{\mathrm{weak}}(p)$ & $0.45(1-p)+0.15p$ \\
& Sample-weight multiplier / cap $\kappa_w$, $w_{\max}$ & $4$ / $32$ \\
\bottomrule
\end{tabularx}
\end{table*}

Table~\ref{tab:G1} gives the defaults used unless an experiment specifies a variant. Embeddings and logits satisfy the listed hard-equivalence tolerance at initialization and on trained all-KEEP trajectories.

\subsection{Data, Scoring, and Statistical Procedures}
\label{app:G2}

\begin{table*}[!tbp]
\centering
\caption{Readout-pathway ablations under rule-based allocation (MSC-derived 535, frozen Qwen2.5-7B, $k=8$). All rows share the main model parameters. Per-question position counts are identical within each ratio. F1, EM, NLL, and budget are owner-weighted. Wrong$\to$right / right$\to$wrong count questions whose EM changes after disabling a pathway. $\Delta$EM is the question-weighted gain from enabling the pathway. Intervals use paired owner bootstrap. Bold marks adjusted $p<0.05$. Exact $p$ values are Bonferroni-adjusted over eight comparisons (Section~\ref{sec:setup}).}
\label{tab:G2}
\footnotesize
\setlength{\tabcolsep}{3pt}
\begin{tabular}{lcccccccc}
\toprule
\makecell{Rule target\\ratio $\rho$} & $R_{\mathrm{all}}$ & Readout & F1 & EM & NLL & \makecell{Wrong$\to$right /\\right$\to$wrong} & $\Delta$EM [95\% interval] & \makecell{Adjusted\\$p$} \\
\midrule
$0.91$ & $0.909$ & All enabled & 95.8 & 60.1 & 0.182 & Reference & \textemdash & \textemdash \\
$0.91$ & $0.909$ & No Reader & 95.0 & 58.2 & 0.224 & $4$ / $14$ & $+1.87$[$0.56, 3.36$] & 0.1702 \\
$0.91$ & $0.909$ & No Global & 95.0 & 58.2 & 0.234 & $5$ / $15$ & $+1.87$[$0.37, 3.38$] & 0.2471 \\
$0.85$ & $0.826$ & All enabled & 91.1 & 41.6 & 0.438 & Reference & \textemdash & \textemdash \\
$0.85$ & $0.826$ & No Reader & 90.7 & 40.5 & 0.565 & $4$ / $10$ & $+1.12$[$-0.19, 2.43$] & 1.0000 \\
$0.85$ & $0.826$ & No Global & 90.5 & 40.5 & 0.588 & $5$ / $11$ & $+1.12$[$-0.19, 2.44$] & 1.0000 \\
$0.75$ & $0.775$ & All enabled & 85.0 & 34.1 & 0.651 & Reference & \textemdash & \textemdash \\
$0.75$ & $0.775$ & No Reader & 84.8 & 32.3 & 0.817 & $1$ / $11$ & $+1.87$[$0.75, 3.18$] & 0.0508 \\
$0.75$ & $0.775$ & No Global & 84.6 & 31.9 & 0.848 & $1$ / $13$ & $+2.24$[$0.93, 3.73$] & \textbf{0.0146} \\
$0.60$ & $0.752$ & All enabled & 80.0 & 22.9 & 0.799 & Reference & \textemdash & \textemdash \\
$0.60$ & $0.752$ & No Reader & 80.1 & 22.9 & 0.982 & $4$ / $4$ & $+0.00$[$-1.12, 0.94$] & 1.0000 \\
$0.60$ & $0.752$ & No Global & 79.9 & 22.9 & 1.017 & $6$ / $6$ & $+0.00$[$-1.31, 1.31$] & 1.0000 \\
\bottomrule
\end{tabular}
\end{table*}

\begin{table*}[!tbp]
\centering
\caption{Readout-pathway ablations under learned allocation (MSC-derived 535). All rows share the main parameters, Controller actions, Bank / Global state trajectories, and per-question budgets, with relative budget $0.936$. Metrics are question-weighted, whereas Table~\ref{tab:1} is owner-weighted. Change counts and $\Delta$EM follow Table~\ref{tab:G2}. Exact owner-level $p$ values are Bonferroni-adjusted over two comparisons.}
\label{tab:G3}
\small
\setlength{\tabcolsep}{5pt}
\begin{tabular}{lcccccc}
\toprule
Readout at inference & F1 & EM & NLL & \makecell{Wrong$\to$right /\\right$\to$wrong} & $\Delta$EM [95\% interval] & Adjusted $p$ \\
\midrule
All enabled & 95.46 & 84.49 & 0.1123 & Reference & \textemdash & \textemdash \\
No Reader & 95.46 & 84.11 & 0.1274 & $2$ / $4$ & $+0.37$[$-0.56, 1.31$] & 1.0000 \\
No Global & 95.25 & 83.36 & 0.1315 & $1$ / $7$ & $+1.12$[$0.19, 2.24$] & 0.1406 \\
\bottomrule
\end{tabular}
\end{table*}

\paragraph{Record construction and filtering.}
Source values are whitespace-normalized and capped at $220$ characters for profiles and $260$ for dialogue. The evaluation loader excludes records longer than $384$ tokens and values longer than $48$ tokens, and excludes histories with fewer than two eligible records. Question targets use the first and last retained record tags, with the latest retained value serving as the answer for each target tag. The resulting MSC-derived development probe contains all $535$ questions from the $268$ retained owners. Under question weighting, one correct answer contributes approximately $0.19$ EM percentage points. Under owner weighting, its contribution depends on the owner's question count.

\paragraph{Lexical scoring and likelihood.}
SQuAD normalization lowercases predictions and gold answers, removes punctuation and the articles \emph{a}, \emph{an}, and \emph{the}, and collapses whitespace~\citep{rajpurkar2016squad}. EM assigns $1$ to an exact normalized-string match. F1 measures bag-of-words overlap. Answer NLL concatenates the prompt and gold answer, runs a teacher-forced forward pass, and averages the negative natural logarithm of each gold token's conditional probability over answer tokens, excluding the end token. Conditional log-probability also underlies generation metrics such as BARTScore~\citep{yuan2021bartscore} and GPTScore~\citep{fu2024gptscore}. Differences, ranges, and budget ratios are calculated from unrounded values before display rounding.

\paragraph{Coverage and local judging.}
LongMemEval Cover applies local rules by question type. Ordinary answers require a normalized substring match or the presence of all answer words. Multipart answers must satisfy every part. Temporal questions with a single integer answer allow a tolerance of $\pm1$. Preference questions require at least half of the answer words longer than two characters, and abstention questions use abstention-phrase rules. Brief requires Cover and $n_{\mathrm{pred}}\le\max(12,3n_{\mathrm{gold}})$, where the two counts denote normalized prediction and gold-answer words. The local Qwen judge uses the official LongMemEval yes/no prompts with Qwen2.5-7B-Instruct. For $500$ questions and accuracy near $0.11$, the binomial standard error of a single accuracy estimate is approximately $1.4$ percentage points.

\paragraph{Paired statistics.}
For owner $u$ among the $268$ owners, let $D_u$ sum the differences in binary EM indicators between full readout and the disabled pathway over that owner's questions. The question-weighted effect in EM percentage points is $\frac{100}{535}\sum_{u=1}^{268}D_u$. The two-sided exact test enumerates sign flips of the $D_u$, assuming independent owners and invariance under swapping each owner's complete paired outcomes under the null~\citep{winkler2014permutation}. Pointwise $95\%$ percentile intervals use $100{,}000$ owner-clustered paired bootstrap resamples with seed $20260923$~\citep{field2007bootstrapping}. The four rule-width ratios and two pathways define one family of eight comparisons. The two learned-trajectory comparisons define a separate family. Significance requires Bonferroni-adjusted exact-test $p<0.05$~\citep{dunn1961multiple}. Bootstrap intervals estimate effect uncertainty without multiplicity adjustment. Complete readout results appear in Tables~\ref{tab:G2} and~\ref{tab:G3}.

\paragraph{Training variants and notation.}
Let $s_{\mathrm{policy}}$ denote completed policy-phase updates. The sweep in Table~\ref{tab:B1} and training in Table~\ref{tab:E4} use $p=\operatorname{clip}(s_{\mathrm{policy}}/5081,0,1)$ and a total target of $384+5081=5465$ updates. Actual counts accompany the corresponding results. The fixed-ratio and retention-rule training arms in Table~\ref{tab:E4} optimize answer NLL plus imitation of rule actions, with forced softening, the target band, and Writer hinge disabled. The complete \texttt{full} curriculum includes forced softening, the target band, Writer hinge, and KEEP-wrong sample weighting. Parameter variants specify $X=v$ alongside \texttt{full}. The hard reference has zero softening. The all-KEEP \texttt{noshrink} path is equivalent to the hard reference. Unless specified otherwise, an em dash denotes an unmeasured or undefined quantity.

\paragraph{External-component protocols.}
Mem0 and A-MEM use public memory components. LLMLingua-2 calls the official \texttt{compress\_prompt} interface once per complete history. Appendix~\ref{app:A} specifies backbones, integration protocols, and state-isolation boundaries. The reconstruction probe depends on identifier lookup, wording, and answer format. For \emph{Stable candidate P2-0: I'm a man.}, questions identify P2-0 and expect its record value. \emph{User is a man.} omits this identifier and changes the normalized answer.

\subsection{Complete Fixed-Model Readout Ablations}
\label{app:G3}

Tables~\ref{tab:G2}--\ref{tab:G3} give the full measurements underlying Table~\ref{tab:6}. Absolute metric values under rule-based allocation are owner-weighted. Those under learned allocation and all paired EM differences are question-weighted. Interventions within each allocation condition share per-question memory positions and model parameters. Global interventions retain state updates and Writer conditioning.

Under rule allocation at $\rho=0.60$, disabling Reader changes four answers from wrong to right and four from right to wrong, yielding zero net EM change. Under learned allocation, Global's pointwise bootstrap interval is $[0.19,2.24]$, while its multiplicity-adjusted exact-test $p$ is $0.1406$. The intervals are unadjusted for multiple comparisons. Significance follows the exact-test procedure in Section~\ref{sec:setup}.

\end{document}